\documentclass[review]{elsarticle}
\graphicspath{ {./figures/} }
\usepackage{hyperref}
\DeclareUnicodeCharacter{00A0}{ }
\usepackage[utf8x]{inputenc}
\usepackage{float}
\usepackage{verbatim} 
\usepackage{apalike}
\restylefloat{figure}
\restylefloat{table}
\usepackage{color}
\usepackage{amssymb}
\usepackage{amsmath}
\usepackage{xcolor}
\usepackage{caption}
\usepackage{subcaption}
\usepackage{multirow}
\usepackage{booktabs}
\usepackage{threeparttable}

\journal{Expert Systems with Applications}

\biboptions{authoryear}

\begin{document}
\begin{frontmatter}
\begin{titlepage}
\begin{center}
\vspace*{1cm}

\textbf{ \large Efficient Neural Architecture Search for Emotion Recognition}

\vspace{1.5cm}
Monu Verma $^{a}$ (monuverma.cv@gmail.com), Murari~Mandal $^{b}$, Satish~Kumar~Reddy $^{c}$, Yashwanth~Reddy~Meedimale $^{d}$, Santosh Kumar Vipparthi$^{e}$\\

\hspace{10pt}

\begin{flushleft}
\small  
$^a$ Electrical and Computer Engineering, University of Miami, Florida, USA,(33146) \\
$^b$ School of Computer Engineering, Kalinga Institute of Industrial Technology, Bhubaneswar, Odisha, India, (751024)\\
$^c$JP Morgan Chase \& Co, Hyderabad, Telangana, India, (500081)\\
$^d$Walmart Global Tech, Hyderabad, Telangana, India, (560103)\\
$^e$CVPR Lab, Dept. of Electrical Engineering, Indian Institute of Technology, Ropar, Roopnagar, India (140001)

\vspace{1cm}
\textbf{Corresponding Author:} \\
Santosh Kumar Vipparthi \\
CVPR Lab, Dept. of Electrical Engineering, Indian Institute of Technology, Ropar, Roopnagar, India (140001) \\
Tel: +91-9549658135 \\
Email:skvipparthi@iitrpr.ac.in

\end{flushleft}        
\end{center}
\end{titlepage}

\title{Efficient Neural Architecture Search for Emotion Recognition}

\author[label1]{Monu Verma}
\ead{monuverma.cv@gmail.com}

\author[label2]{Murari Mandal}
\ead{murari.mandalfcs@kiit.ac.in}
\author[label3]{Satish Kumar Reddy }
\ead{mskr181298@gmail.com}
\author[label4]{Yashwanth Reddy Meedimale}
\ead{yashwanth3130@gmail.com}
\author[label5]{Santosh Kumar Vipparthi \corref{cor1}}
\ead{skvipparthi@iitrpr.ac.in}

\cortext[cor1]{Corresponding author.}
\address[label1]{Electrical and Computer Engineering, University of Miami, Florida, USA,(33146)}
\address[label2]{School of Computer Engineering, Kalinga Institute of Industrial Technology, Bhubaneswar, Odisha, India, (751024)}
\address[label3]{JP Morgan Chase \& Co, Hyderabad, Telangana, India, (500081)}
\address[label4]{Walmart Global Tech, Hyderabad, Telangana, India, (560103)}
\address[label5]{CVPR Lab, Dept. of Electrical Engineering, Indian Institute of Technology, Ropar, Roopnagar, India (140001)}
\begin{abstract}
Automated human emotion recognition from facial expressions is a well-studied problem and still remains a very challenging task. Some efficient or accurate deep learning models have been presented in the literature. However, it is quite difficult to design a model that is both efficient and accurate at the same
time. Moreover, identifying the minute feature variations in facial
regions for both macro and micro-expressions requires expertise
in network design. In this paper, we proposed to search for a highly efficient
and robust neural architecture for both macro and micro-level
facial expression recognition. To the best of our knowledge,
this is the first attempt to design a NAS-based solution for both
macro and micro-expression recognition. We produce lightweight
models with a gradient-based architecture search algorithm. To maintain
consistency between macro and micro-expressions, we utilize
dynamic imaging and convert micro-expression sequences into
a single frame, preserving the spatiotemporal features in the facial
regions. The EmoNAS has evaluated over 13 datasets (7 macro expression
datasets: CK+, DISFA, MUG, ISED, OULU-VIS CASIA,
FER2013, RAF-DB, and 6 micro-expression datasets:
CASME-I, CASME-II, CAS(ME)\^2, SAMM, SMIC, MEGC2019
challenge). The proposed models outperform the existing state-of-the-art methods and perform very well in terms of speed and
space complexity.
\end{abstract}

\begin{keyword}
Human emotion, micro-expression, macro-expression, neural architecture search (NAS), deep learning
\end{keyword}

\end{frontmatter}

\section{Introduction}
\label{introduction}

Human emotion recognition from visual content has attracted significant attention in the last few decades. Facial appearances are one of the most discriminative features of emotional responses \cite{ekman1971constants}. Therefore, researchers have primarily focused on developing robust facial expression recognition (FER) systems for emotion recognition. It has numerous applications in human-computer interaction, behavior profiling, and smart healthcare solutions.\par

\begin{figure*}[t]
\centering
	\includegraphics[width=1\textwidth ]{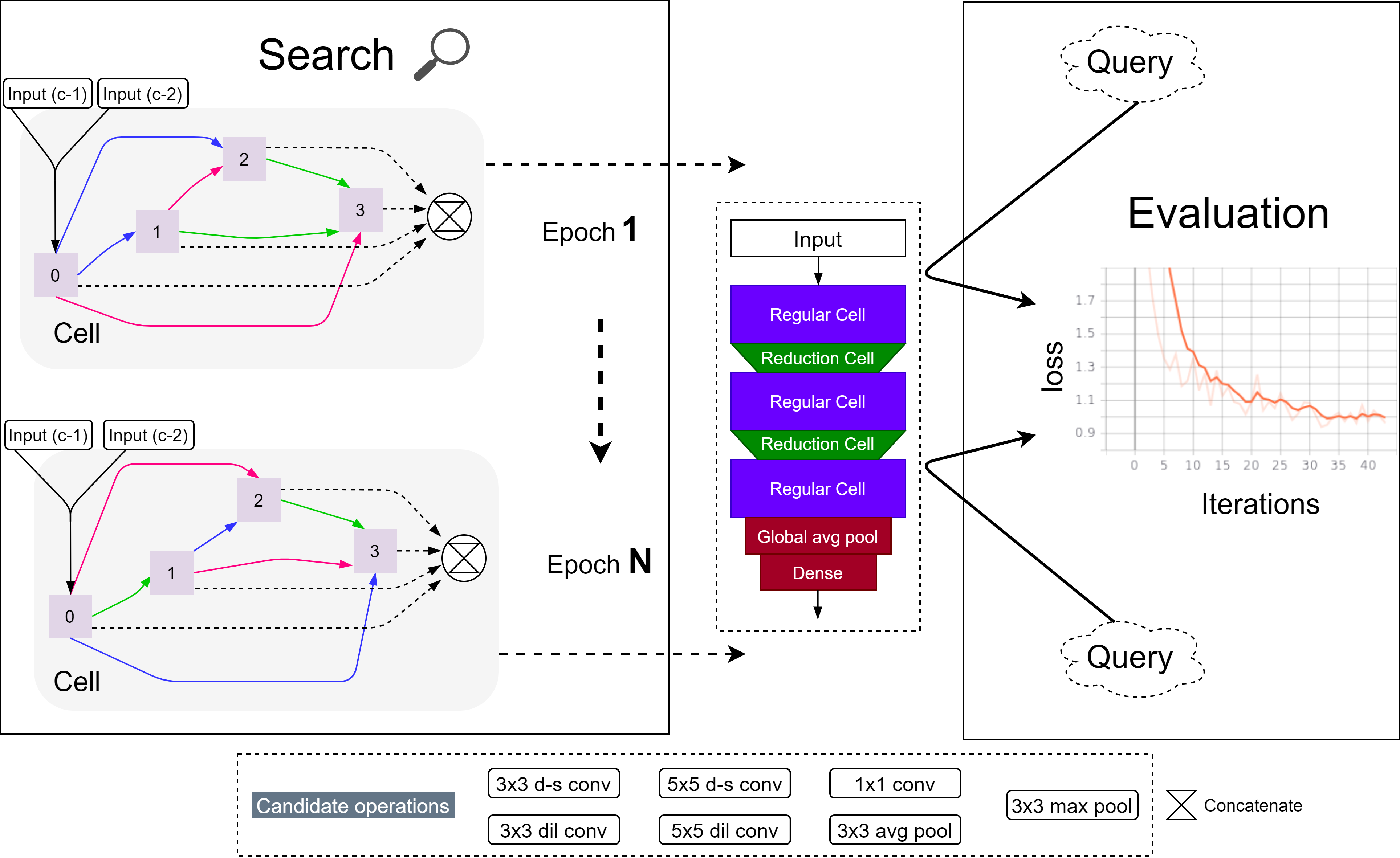}
	\caption{The training and evaluation procedure of the proposed EmoNAS. In the training phase, the regular and reduction cells are optimized. In evaluation, the optimized cells are stacked to classify the input image. \textit{d-s: depthwise-separable, dil: dilated}}
	\label{fig:fig_framework}
\end{figure*}

Facial expressions can be classified into two types: macro and micro-expressions. Macro expressions are considered as a prevailing display of emotions and usually last up to 3 to 4 minutes\cite{ekman2003darwin}. Whereas, micro-expressions are generated when someone tries to hide their actual feelings. Micro-expression appears on the facial region for a fraction of time (1/3-1/22) seconds\cite{verma2019learnet}. The existing state-of-the-art convolutional neural network (CNN) based methods have primarily improved upon the FER performance and few others have focused on improving the efficiency through smaller network design \cite{verma2019hinet}. However, state-of-the-art approaches require the substantial effort of human expertise to design an effective neural network architecture. Due to the human efforts involved in deep network design, there is room for improvement in both efficiency and accuracy. Moreover, the subtle variations in certain facial regions lead to changes in emotion class. There is a need for developing robust and lightweight FER models for real-world applications.\par

Recently, there has been a growing interest in developing algorithmic solutions to automatically design an architecture for deep learning models. These algorithms are known as neural architecture search (NAS). Inspired by the differentiable architecture search algorithm \cite{liu2018darts}, in this paper, we present EmoNAS to search for the most robust and efficient neural architecture for FER. The training and evaluation procedure of the proposed EmoNAS is shown in Figure \ref{fig:fig_framework}. As discussed earlier, macro and micro expressions are different in nature, so it is challenging to design a network that can efficiently work for both types of expressions. (more detailed differences between macro and micro expressions are discussed in the supplementary document). Therefore, solutions for both MaEs and MEs are non-identical in the literature.  Therefore, Most of the existing FER algorithms are either designed for macro or micro expression classification. Some of the concepts utilize the macro adaptive features for micro-expression classification \cite{suncao-taf, yang2021merta}. However, there is no single algorithm to solve macro and micro-expression recognition. We made the first attempt to introduce a universal NAS algorithm to solve the generic macro facial expressions as well as micro facial expressions. Usually, a single instance of an image is sufficient for macro expression recognition analysis whereas, MER requires spatiotemporal data due to its fleeting and short-lived nature. Therefore, to maintain a uniform experimental setting, we adopted the dynamic imaging concept to extract a single instance feature map consisting of the spatiotemporal features from the temporal sequence of frames. Therefore, this paper aims to provide a uniform search and training framework to target the challenges in FER and MER using an EmoNAS.  Our proposed framework EmoNAS has the following contributions:

\begin{enumerate}
\item We propose a differentiable architecture search-based algorithm named EmoNAS to handle the challenges in macro and micro-expression with the optimized CNN models. We propose to employ a uniform number of cells in the architecture design as used in the search phase. Moreover, the search space is restricted to a shallower structure to obtain lightweight models for real-time applications.
\item The EmoNAS achieves remarkable efficiency improvement over the existing deep learning models in terms of the final model computational complexity.
\item {The effectiveness of EmoNAS is validated on 13 datasets: 7 macro facial expression recognition (FER) and 6 micro-expression recognition (MER) datasets. It achieves highly competitive results and outperforms {most of all the existing} state-of-the-art methods in these 13 datasets.  The impact of selecting a different number of cells and nodes is extensively studied by conducting 9 ablation experiments on CK+, DISFSA, CASME-I, and CAS(ME)\^2 datasets.}
\end{enumerate}

\section{Related Work}
\subsection{Macro-expression}
Macro-expression recognition, commonly referred to as facial expression recognition (FER), has been widely studied in the literature. The deep learning algorithms for FER~\cite{pons2017supervised,xie2020adversarial} have far outperformed the traditional hard-crafted approaches~\cite{Mandal2019RegionalAA,rivera-2020}. Researchers have developed numerous CNN \cite{verma2019hinet,zhang2016deep} for FER. Fan et al.~\cite{fan2020facial} proposed a two-stage attention network to detect posed and spontaneous expressions. Similarly, Li et al.~\cite{li2020facial} used attention at the patch and whole-face level for effective FER. Li and Deng~\cite{lideng-taf} conducted an extensive experiment to delve into the bias across different FER datasets and also explore the intrinsic causes of the dataset discrepancy. Wang et al.~\cite{Wang_2020_CVPR} studied the effect of ambiguous annotations, low-quality expression images, and obscure facial expressions over deep learning networks. A more detailed study of the deep learning-based FER methods is given in~\cite{li2020deep}.

\subsection{Micro-expression} 
Micro-expression recognition (MER) has received much attention in past few years. The earlier MER works focused on the spatiotemporal feature descriptors and optical flow \cite{sparseMDMO,XuDynamicMap,xia2020revealing}. The recent works have used the CNN, RNN, GANs, 3D-CNN and other deep learning techniques to improve the performance~\cite{monu-ijcnn,suncao-taf,li2020joint,monu-multimedia,wang2020micro,van2019capsulenet,liong2019shallow}.  Liong et al.~\cite{liong2019shallow} proposed a triple stream CNN by using three optical flow features (optical strain, horizontal and vertical optical flow) computed between the onset and apex frames of each sequence. Furthermore, Wang et al.~\cite{wang2020micro} proposed an attention-based residual network to guide the CNN towards the micro-expression regions. More recently, a composite database for MER was presented along with the benchmark experimental evaluation protocol in~\cite{zhang-tkde}. Subsequently, Xia et al.~\cite{xia2020revealing} studied the influence of learning complexity on the composite database. They conclude that the low-resolution input and shallower architecture are more suitable for composite-dataset problems in comparison to the deep architectures. A more detailed study of various MER methods can be found in~\cite{zhou2020survey}. 

\subsection{Neural Architecture Search} Recently NAS algorithms intended for automatically searching and designing CNN architectures have achieved very competitive performance in computer vision tasks such as image classification \cite{real2019regularized} and object detection \cite{zoph2018learning}. Despite their remarkable performance, the earlier NAS algorithms were computationally expensive and demands huge memory footprints. The reinforcement learning (RL)\cite{zoph2018learning} NAS requires 2000 GPU days to get existing architecture for CIFAR-10 and ImageNet. Similarly, another NAS algorithm \cite{real2019regularized} needs 3150 GPU days of evaluation. Furthermore, various NAS algorithms \cite{Bender2018UnderstandingAS, baker2017accelerating, liu2018progressive} have been introduced to speed up the search process. The RL, evolution, MCTS \cite{negrinho2017deeparchitect}, SMBO \cite{liu2018progressive} based NAS algorithms consider the architecture search as a black-box optimization problem over a discrete space, which leads to a huge number of architecture assessments. Therefore,\cite{liu2018darts} relaxed the search space to be continuous to optimize the architecture through gradient descent with respect to its validation set performance. This is the most effective solution to accelerate the search process. Therefore, we use this optimization technique in our work. Recently Yu et al. \cite{yu2020fas,yu2020auto,yu2020searching} proposed three different NAS based algorithm for facial anti-spoofing.The darts-based approaches have focused on searching the inner cell structure, only. Li et al \cite{li2021auto} introduced a NAS-based algorithm Auto-FERNet to automatically search a CNN model on a macro expression dataset. \par

The network design process for the existing deep learning FER and MER methods requires a lot of manual effort. Usually, the designed network performs well over a few expression datasets but does not transfer well to other expression datasets. Moreover, the macro- and micro-expression recognition tasks require quite different algorithmic approaches to obtain robust performance. Thus, it is difficult to design a single network to effectively solve both the FER and MER tasks. The NAS algorithms offer an alternative approach to universally solving both the FER and MER problems by automatically searching for efficient and robust architectures. Motivated by this, we propose the EmoNAS algorithm and experimentally show its effectiveness in FER and MER. To the best of our knowledge, this is the first attempt to solve both macro- and micro-expression tasks with neural architecture search.
\begin{figure}[t!]
\centering
	\includegraphics[width=0.95\textwidth ]{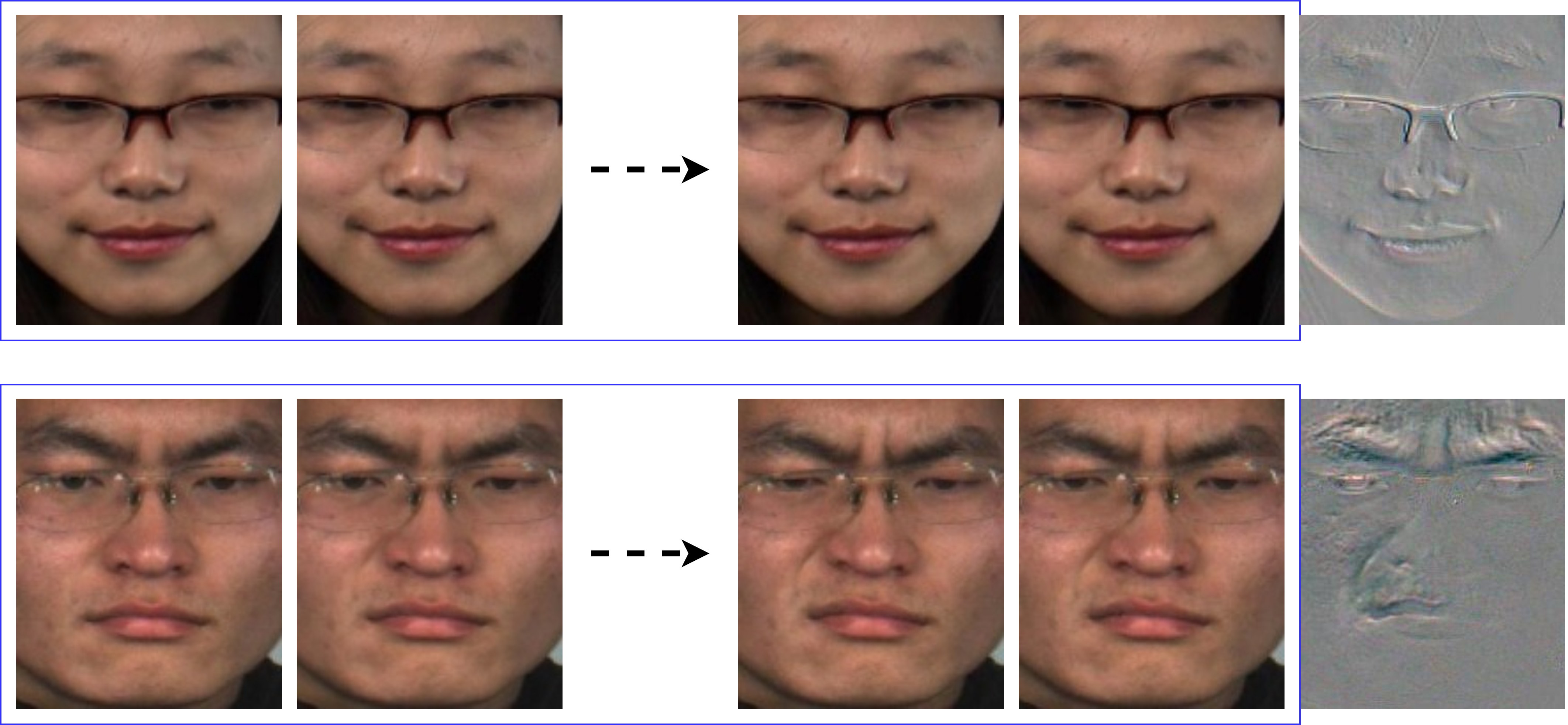}
	\hfill\small \center \hspace{2cm} (a) Video sequence \hspace{1cm} (b) Dynamic image
	\caption{Sample visual representation of the dynamic image response for micro-expression sequences. The dynamic image captures the spatiotemporal features for subtle changes in facial regions.}
	\label{fig:fig_dyn}
\end{figure}
\section{Dynamic Imaging}
The main aim of this paper is to provide a uniform search and training framework to address the challenges in FER and MER using a NAS method. Usually, a single instance of the image if sufficient for FER analysis whereas, MER requires spatiotemporal data due to its fleeting and short-lived nature. Therefore, to maintain a uniform experimental setting, we adopted the dynamic imaging concept to extract a single instance feature map consisting of the spatiotemporal features from the temporal sequence of frames. Most of the existing MER approaches~\cite{van2019capsulenet, li2020joint, ststnet, liu2019neural} rely only on the apex frame for the analysis. However, some studies emphasize the importance of dynamic aspects for detecting the subtle changes~\cite{ambadar2005deciphering} and their effect on the performance of MER. In a micro-expression (MEs) video, each frame has its own significance in the identification of the emotion class. Some recent approaches~\cite{ReddyFuse, Li3D-Flow} utilized video sequences to design an effective MER system. However, all publicly available MER datasets hold videos with a variant number of frames. In order to normalize the video sequences, some approaches have utilized time interpolation algorithms. This may lose or alter the domain knowledge of micro-expressions by shearing or filling holes in between the frames. To overcome the above-mentioned issues and embed the subtle facial changes, we have utilized the dynamic imaging concept~\cite{bilen2016dynamic} to generate a single instance of an image. Dynamic imaging represents the video information into a single instance by conserving high stake active dynamics of MEs. It also ensures uniform search and training architecture for both macro and micro-expression recognition.\par

The dynamic imaging technique has been effectively used in the recent literature for micro-expression recognition~\cite{verma2019learnet}. The dynamic image interprets the content of the video by aiming at the facial moving regions and compresses that into a single instance. The dynamic image is an RGB response holding the spatiotemporal features of a video sequence. The dynamic image \(d^*\) is calculated by Eq. 1-Eq. 5.

\begin{equation}\label{eq0_1}
    d^*=\sum_{p=1}^{N}{\mathbb{Z}_{p}}
\end{equation}
where $\mathbb{Z}_{p}$ and \(N\in q\) represent the $p^{th}$ intermediate motion image and a total number of frames in a video \(V\), respectively. The intermediate motion image is calculated by using Eq. 2. 
\begin{equation}
\label{eq0_2}
    {\mathbb{Z}_{p}}=V_{p}\times F(p)
\end{equation}
where $F(p)$ implies the frame weight and is calculated by using Eq. 3.
\begin{equation}
\label{eq0_3}
     F(p)=\sum_{l=p}^{q}\mathbb{R}(1,l)
\end{equation}
where \(V_{p}\in V\) represents the $p^{th}$ frame of the video V, \(p=1,2,...,q\) and \(\mathbb{R}\) is a ranking function which upgrade the rank of frames as follows:
\begin{align}
    \mathbb{R}(1,l)=\frac{2\times \mathbb{I}(1,l)-q}{\mathbb{I}(1,l)}\\
    \mathbb{I}(1,l)=(l,l+1,l+2,...q)
\end{align}

where \(q\) implies the total frames in video sequences. $l$ is a index of the frames in range $\left[1,q\right]$. We have shown sample dynamic responses in Figure \ref{fig:fig_dyn}. It can be observed that both the uniform and non-uniform features of a video stream are embedded in a single frame. We use these dynamic images to search and train the NAS models for micro-expression recognition.

\section{EmoNAS}
\begin{figure}[t!]
\centering
	\includegraphics[width=0.95\textwidth ]{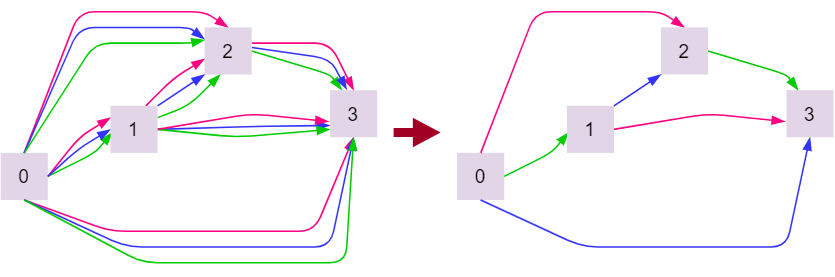}
	\caption{An overview of the cell architecture. Initially, the operations on the edges are unknown. The operations are selected from 7 candidate operations.}
	\label{fig:fig_cell}
\end{figure}

The proposed EmoNAS algorithm for architecture search is depicted in Figure~\ref{fig:fig_framework}. We search for a computationally viable cell as the building block of the final architecture. The cell structure is shown in Figure~\ref{fig:fig_cell}. The objective is to discover the optimal cell structure that can lead to the best architecture design for the specific dataset. We use two types of cells: regular cells and reduction cells. Regular cells carry the feature maps to the next cell by maintaining spatial resolution, while reduction cells imply the down-sampling. Further, the learned cell is stacked to form a shallower convolutional network.  {In the existing method~\cite{liu2018darts}, the structures discovered through the search process are not necessarily optimal for FER/MER evaluation due to the deep dense final architecture. In literature~\cite{verma2019learnet, pasupa2016comparison}, it is well proven that deep networks fail to achieve adequate performance over small-sized datasets. Most of the existing benchmark FER datasets have a lesser number of image/video samples. Also, the features extracted in the search step might get lost or distorted by more cell stacking while training due to repetitive cross-correlation and pooling operations. We are also motivated by the fact that the properties of deep and shallow networks are quite different which has been studied in the literature~\cite{srivastava2015training}. Therefore, we stacked search cells (5 cells only) within the premise of shallower networks to ensure better efficiency for real-time applications. Through EmoNAS, we obtain lightweight networks with a minimum number of cells without compromising the quantitative performance.}

\subsection {Architecture Search Space and Optimization}
Our goal is to discover a robust cell topology and use it to develop a convolutional network of cells. A cell is characterized as a directed acyclic graph having $N$ nodes as shown in Figure~\ref{fig:fig_cell}. The nodes represent the feature maps in convolutional networks and the directed edge $(m, n)$ represents an operation $o^{(m,n)}$ that transforms input $x^{(m)}$ into response feature maps. Each cell has 2 input nodes and 1 output node. The cell outputs from the previous layers are used as input nodes. Eventually, the cell structure is formed by learning the set of operations on its edges. A node $k$ in a cell $c$ can be defined by using 5-tuples \((P_{1},P_{2},Q_{1},Q_{2},S)\) where, \(P_{1}, P_{2}\in \mathbb{P}_{k}^{c}\),  \(Q_{1}, Q_{2} \in \mathbb{Q}\) and $S$ represent selection of input tensors, selection of employed layers over selected input tensors and method used to form an output tensor \(OT_{l}^{C}\). Further, the output tensor \(OT^{C}\) of a cell can be calculated by using Eq. \ref{eq1_1}.

\begin{equation}
OT^{C} = OT_{1}^{C}||OT_{2}^{C}||...||OT_{B}^{C}
\label{eq1_1}
\end{equation}
where \(||\) indicate the concatenation operation.\par

The set of candidate operations are defined by $\kappa$, where each operation is denoted by $o(\cdot)$. The continuous search space is  used by relaxing the choice of a particular operation to a softmax over all possible operations as given in Eq. \ref{eq2}.

\begin{equation}
\Tilde{o}^{(m,n)}(x)=\sum_{o\in \kappa} \frac{exp(\alpha^{(m,n)}_o)}{\sum_{{o}'\in \kappa}exp(\alpha_{o'}^{(m,n)})}o(x) \label{eq2}
\end{equation}

where the operation mixing weights for a pair of nodes $(m,n)$ are learned by a vector $\alpha^{(m,n)}$ of dimension $|\kappa|$. After the completion of search, a discrete architecture is retrieved by replacing each mixed operation with the most likely operation. We leverage the approximation scheme given in~\cite{liu2018darts} to perform the optimization.\par
Let $\lambda_{train}$ and $\lambda_{val}$ represent the training and the validation loss, respectively. These losses are calculated based on both the architecture $\alpha$ and weights $w$ in the network. The goal for EmoNAS is to discover~$\alpha^*$ that minimizes the validation loss $\lambda_{val}(w^*,\alpha^*)$. The weights $w^*$ allied with the architecture are computed by minimizing the training loss $w∗ = argmin_w\;\lambda_{train} (w,\alpha^*)$. 
Thus, the bilevel optimization problem with $\alpha$ as the upper-level and $w$ as the lower-level variable is represented in Eq. \ref{eq3} and Eq. \ref{eq4}.
\begin{equation}
\min_\alpha\;\; \lambda_{val}(w^*(\alpha),\alpha)\label{eq3}
\end{equation}
\begin{equation}
s.t. \;\;\;\;w^*(\alpha) = argmin_w\;\;\lambda_{train}(w,\alpha)\label{eq4}
\end{equation}
The detailed procedure and approximation technique is outlined in \cite{liu2018darts}.

\subsection{Network Configurations}
The candidate operation space $\kappa$ consists of the following set of 7 operations: $3\times3$ depthwise-separable convolution, $5\times5$ depthwise-separable convolution, identity, $3\times3$ average pooling, $3\times3$ max pooling, $3\times3$ dilated convolution, $5\times5$ dilated convolution and skip. connection. The network search and evaluation process is discussed below.

\subsubsection{Architecture Search} 
We use 5 cells with each containing 7 nodes. We intentionally select lower number of cells to ensure discovery of lightweight and efficient networks. Similarly, the search space is also limited to only 7 operations. These design decisions helped us to create very lightweight architectures for FER over different datasets. We were also motivated by the success of shallower CNN models in FER to search the shallower architectures through NAS. To compose the nodes in the discrete architecture, we retain the top-2 strongest operations (from distinct nodes) among all the candidate operations. The results analysis discussed in Section~\ref{exp} validate that search space with 5 cells achieve better performance as compared to DARTS with significant performance improvement over all the FER datasets. 

\subsubsection{Architecture Evaluation}
In order to evaluate the discovered cell, we stack same 5 copies of the cells (searched by architecture search), but untied weights with using either stride 1 or stride 2, as shown in Figure \ref{fig:fig_framework} to generate a network. The network is generated by stacking reduction cells at 1/3 and 2/3 locations, the rest of locations are stacked with regular cell in proposed EmoNAS. The number of cells can be adjusted to improve the accuracy/efficiency of the network. We analyze the impact of different parameter changes through ablation study in Section~\ref{ablation}. Each cell consists of two input nodes and one output node, where input nodes are previous cell outputs and output node is concatenation on all intermediate nodes. At the end of the network, we use global average pooling, followed by a softmax classification layer. Furthermore, we trained the model developed by stacking the cells, on the relevant dataset. {More implementation details are presented in Section \ref{impl}}.
\begin{figure}[t!]
\centering
	\includegraphics[width=0.9\textwidth, height=5in]{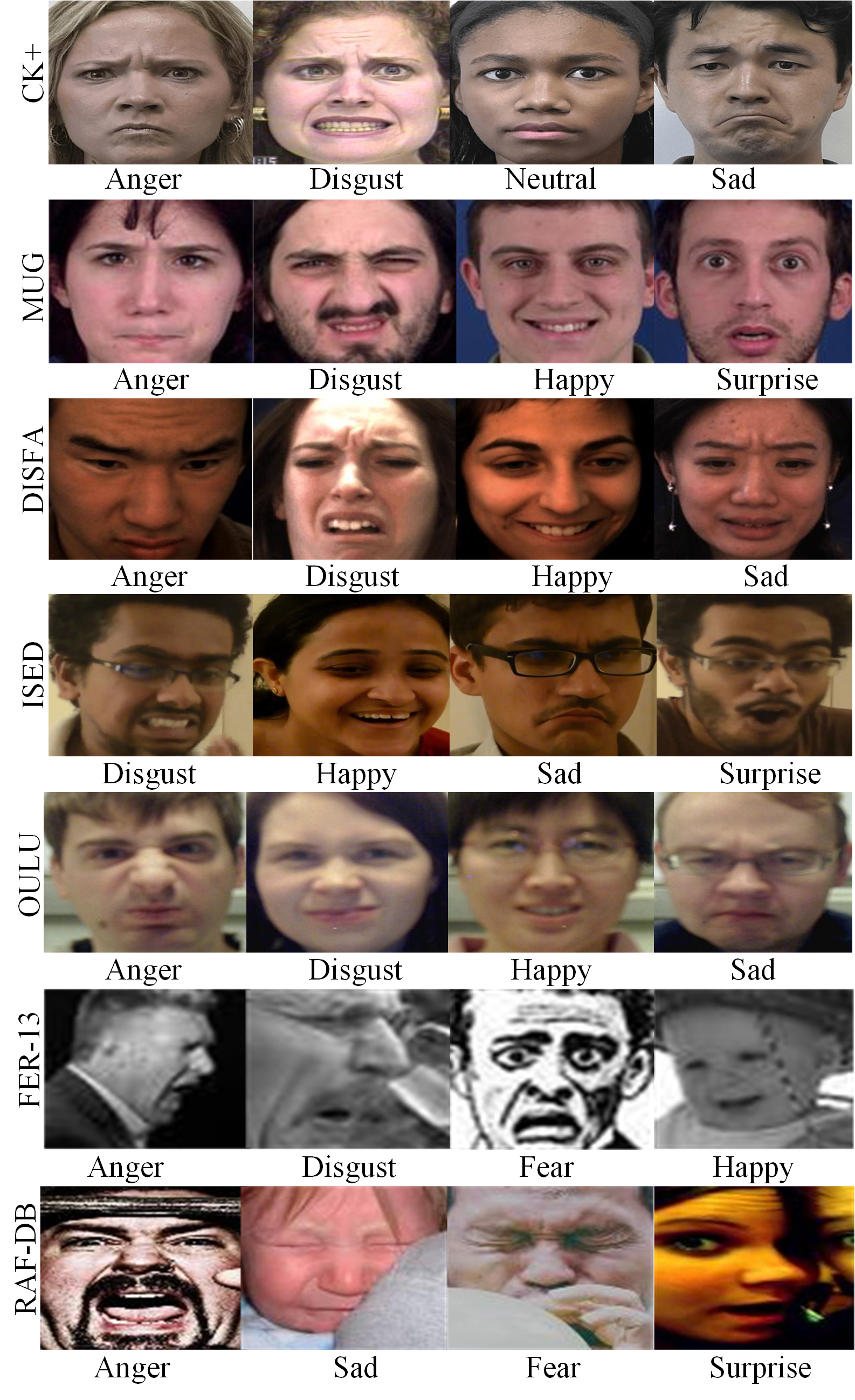}
	\caption{Illustration of different expression samples of macro datasets with challenging tasks a) CK+: posed expressions, b) MUG: posed expressions with enriched resolution, c) DISFA: spontaneous expressions, d) ISED: spontaneous expressions, e) OULU: illumination variations, f) FER-13: Real-time wild expressions, and g) RAF-DB: in-the-wild expressions.}
	\label{fig:dataset_samples}
\end{figure}
\begin{figure*}[t!]
\centering
\begin{subfigure}{.45\linewidth}
    \centering
    \includegraphics[width=1\textwidth]{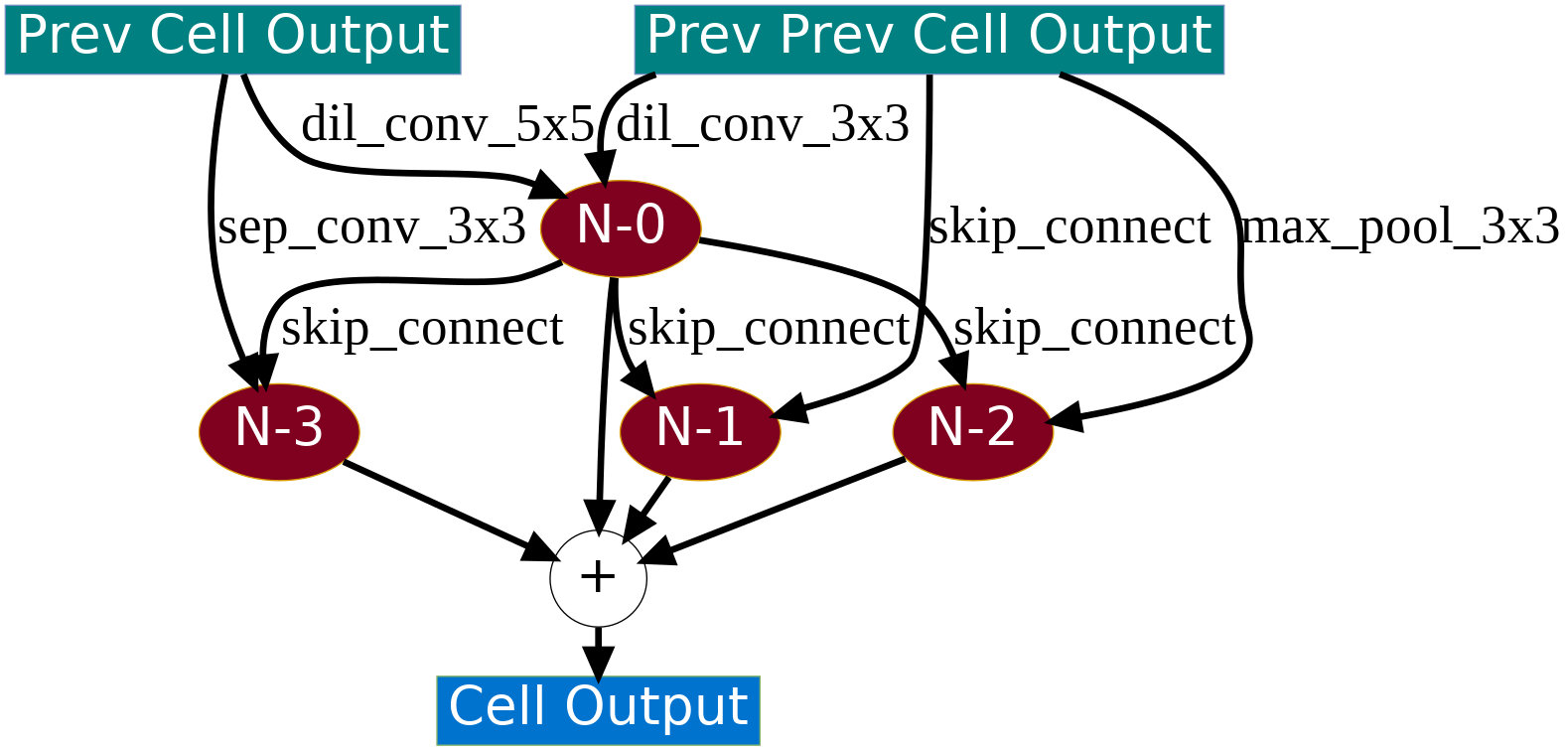}
    \caption{}\label{fig:6a}
    \hfill
\end{subfigure}
\begin{subfigure}{.5\linewidth}
    \centering
    \includegraphics[width=1\textwidth]{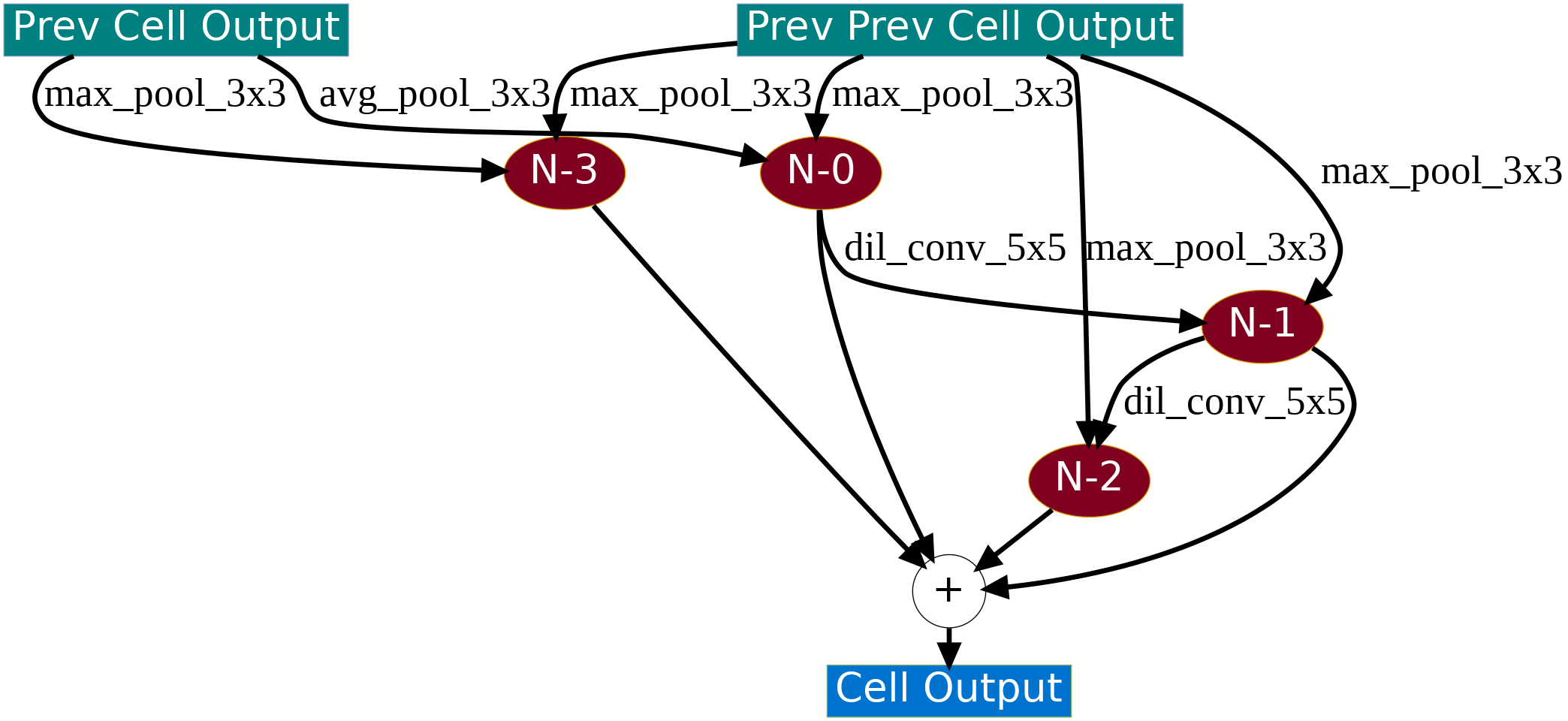}
    \caption{}\label{fig:6b}
    \hfill
\end{subfigure}

\begin{subfigure}{0.45\linewidth}
    \centering
    \includegraphics[width=1\textwidth]{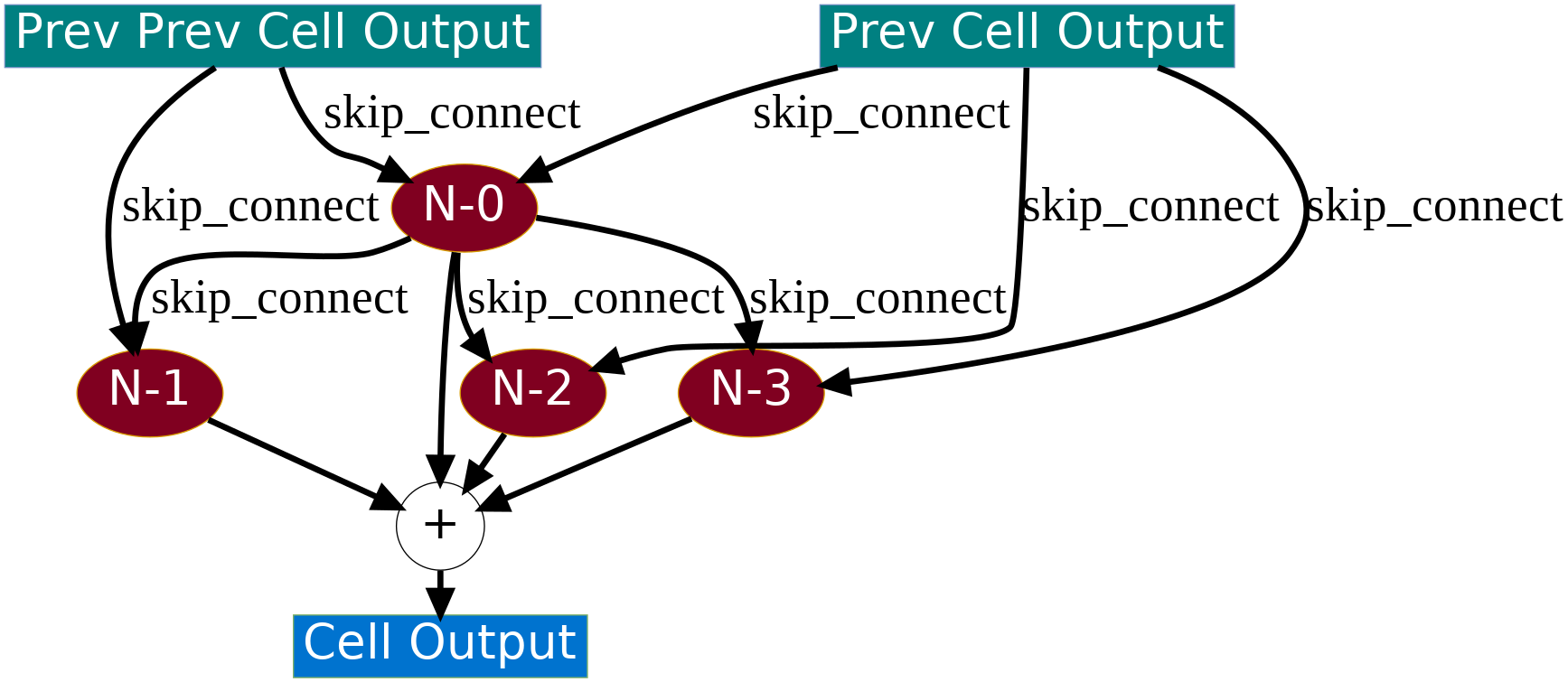}
    \caption{}\label{fig:6e}
\end{subfigure}
\begin{subfigure}{0.5\linewidth}
  \centering
  \includegraphics[width=1\textwidth]{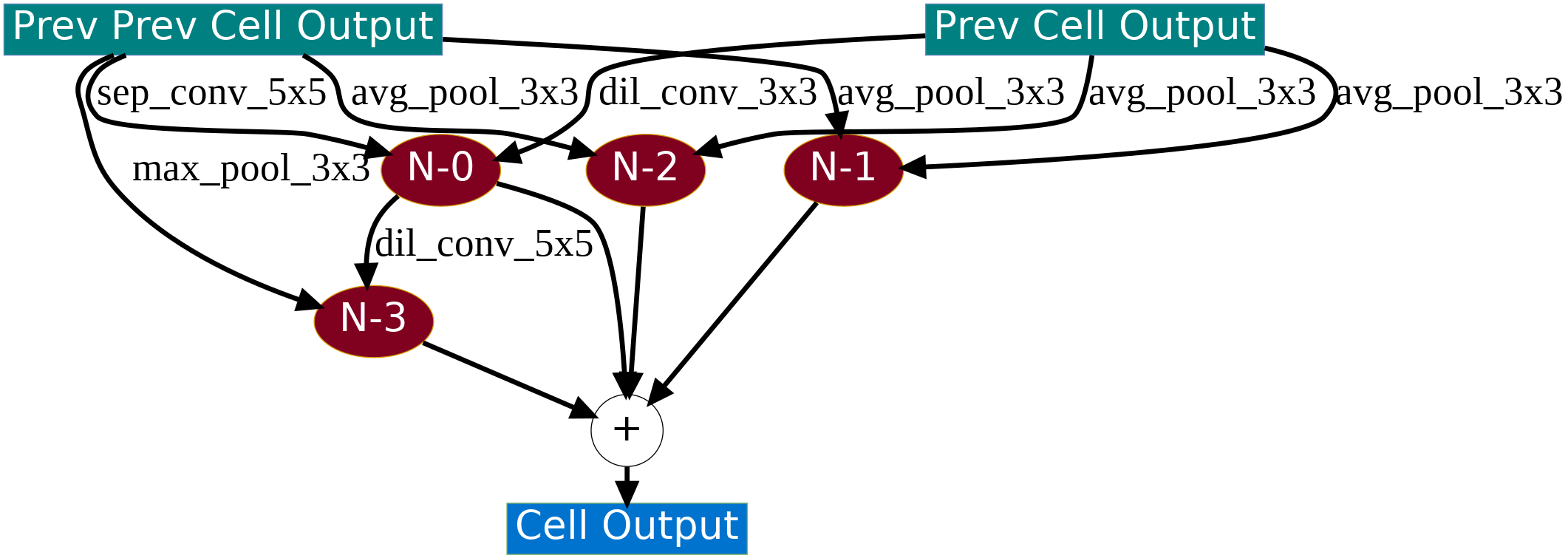}
  \caption{}\label{fig:6f}
\end{subfigure}
\caption{The final regular and reduction cell structures discovered by EmoNAS on (a,b) CK+ (6-Exp), and (c,d) MUG (6-Exp).}
\label{fig:fig_ck_dis_mug}
\end{figure*}

\section{Experiments and Results}
\label{exp}
\subsection{Experimental Details}
This section discusses the implementation details, dataset, and evaluation strategies.
Further, a comparative study of the proposed EmoNAS
and state-of-the-art approaches is presented. We carry
out the ablation study and computational analysis in the following
subsection. 
document.
\subsubsection{Implementation Details}\label{impl}
 {Initially, to search for the best possible cell structure, we used 50 epochs and 16 for batch size due to GPU memory constraints. The main goal of the cell search is to learn the best hyper-parameter $\alpha$ that defines the best possible operations in the cell structure. We used stochastic gradient descent (SGD) optimizer with a momentum of 0.9, a cosine
learning rate of 0.007, and a weight decay of $3e^{-4}$.
Further, we stacked 5 sets of discovered cells with 7
hidden nodes to design the final EmoNAS architecture. For training a model, similar settings of
searching like SGD optimizer with an initial learning rate
of 0.007, weight decay $3e^{-4}$, and momentum 0.9 are initialized.
The batch size is set to 16, and epochs are set to
70 for training a model. The cross-entropy loss function is
used for loss optimization. We implement our model with Pytorch 0.3.1 and run all experiments on an NVIDIA RTX 2080Ti GPU. For each tested dataset, the architecture search is conducted. The same NAS code and hyper-parameters (different image sizes) are used for all the datasets.} 

\begin{table}[t!]
\footnotesize
    \centering
    \caption{Comparative results of the proposed EmoNAS and existing state-of-the-art FER methods on CK+ dataset. }
    \label{tab:tab_ck+}
    \begin{threeparttable}
 \begin{tabular}{l l c c c}
        \toprule
        & & \multicolumn{2}{c}{\textbf{\textit{CK+}}}\\
        \cmidrule(r){3-4}
        \textbf{\textit{Method}} & \textbf{\textit{Pub-Yr}} & {\textit{6 Exp}} & {\textit{7 Exp}}\\
        \midrule
        RADAP \cite{Mandal2019RegionalAA} & IP-19 &88.48 & 83.72\\
        sLSP+LB \cite{rivera-2020} & TAFF-20& 96.77 & 95.13\\
        VGG16 {$^{12}$} \cite{Simonyan2014VeryDC}& Arxiv-15 & 91.31 & 88.18\\
        VGG19 {$^{12}$} \cite{Simonyan2014VeryDC}& Arxiv-15 & 89.98 & 78.71\\
        ResNet50 {$^{12}$} \cite{He2015DeepRL}& CVPR-16 & 89.32 & 87.31\\
        HiNet \cite{verma2019hinet}& LCS-19 & 91.40 & 88.60\\
        DCMA-CNN \cite{Xie2019FacialER}& TMM-19 & 93.46 & N/A\\
        DLP-CNN \cite{li2017reliable}& CVPR-17 & 95.78 & N/A\\
        Lopes \cite{Lopes2017FacialER}& PR-17 & 96.76 & 95.75\\
        IA-gen \cite{Yang2018IdentityAdaptiveFE}& FG-18 &96.57 & N/A\\
        Khor-Net \cite{Khorrami2015DoDN}& ICCVW-15 & 95.70 & N/A\\
        IF-GAN \cite{Cai2019IdentityFreeFE}& Arxiv-19 &95.90 & N/A\\
        DARTS {*} \cite{liu2018darts}& ICML-19 &91.50 & 95.01\\
        P-DART {*} \cite{chen2019progressive}& ICCV-19 &85.93 &86.85 \\
        Auto-FERNet \cite{li2021auto}& TNSE-21 &N/A&{98.89}\\
        ViT-SE \cite{aouayeb2021learning}& Arxiv-21 &N/A&\textbf{99.80}\\
        Squeeze ViT \cite{kim2022facial}& Sensors-21 &N/A&{99.54}\\
        
        \midrule
        \textbf{EmoNAS} & \textbf{Ours} &\textbf{97.13} & \textbf{96.30}\\
        \bottomrule
    \end{tabular}
    \begin{tablenotes}[para,flushleft]
   \textit{ {Here, * and $A^{a}[b]$ implies for re-implemented results and a: results taken from, b: reference of the method. While, Pub-Yr, 6 Exp, and 7 Exp represent Publication-Year, 6 expression, and 7 expression classes.}
   }
  \end{tablenotes}
  \end{threeparttable}
\end{table}

\begin{figure*}[t!]
\centering

\begin{subfigure}{0.48\linewidth}
    \centering
    \includegraphics[width=1\textwidth, height=1.4in]{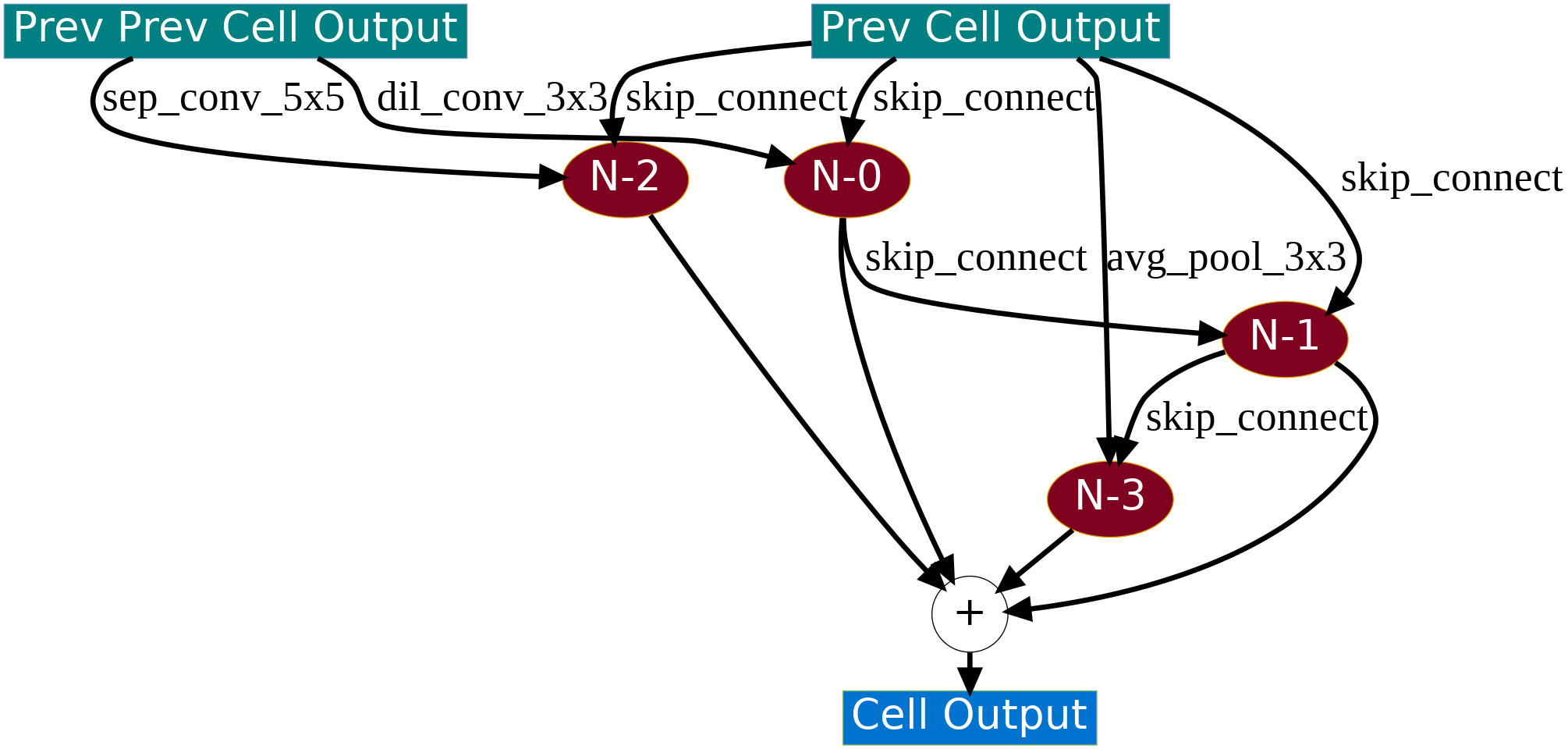}
    \hfill
    \caption{}
    \label{fig:7c}
    \hfill
\end{subfigure}
\hfill
\begin{subfigure}{0.48\linewidth}
  \centering
  \includegraphics[width=1\textwidth, height=1.3in]{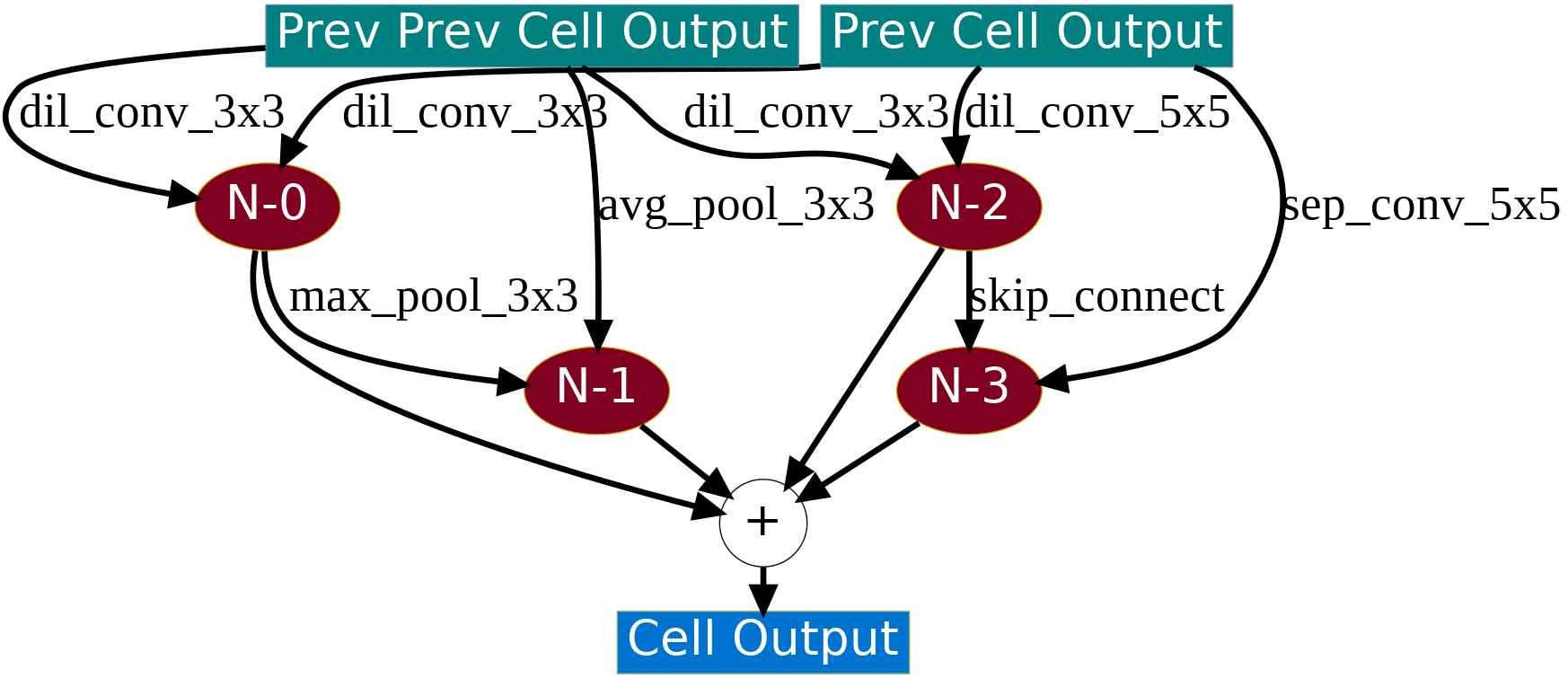}
  \hfill
  \caption{}\label{fig:7d}
\end{subfigure}
\bigskip

\begin{subfigure}{0.48\linewidth}
  \centering
  \includegraphics[width=1\textwidth, height=1.35in]{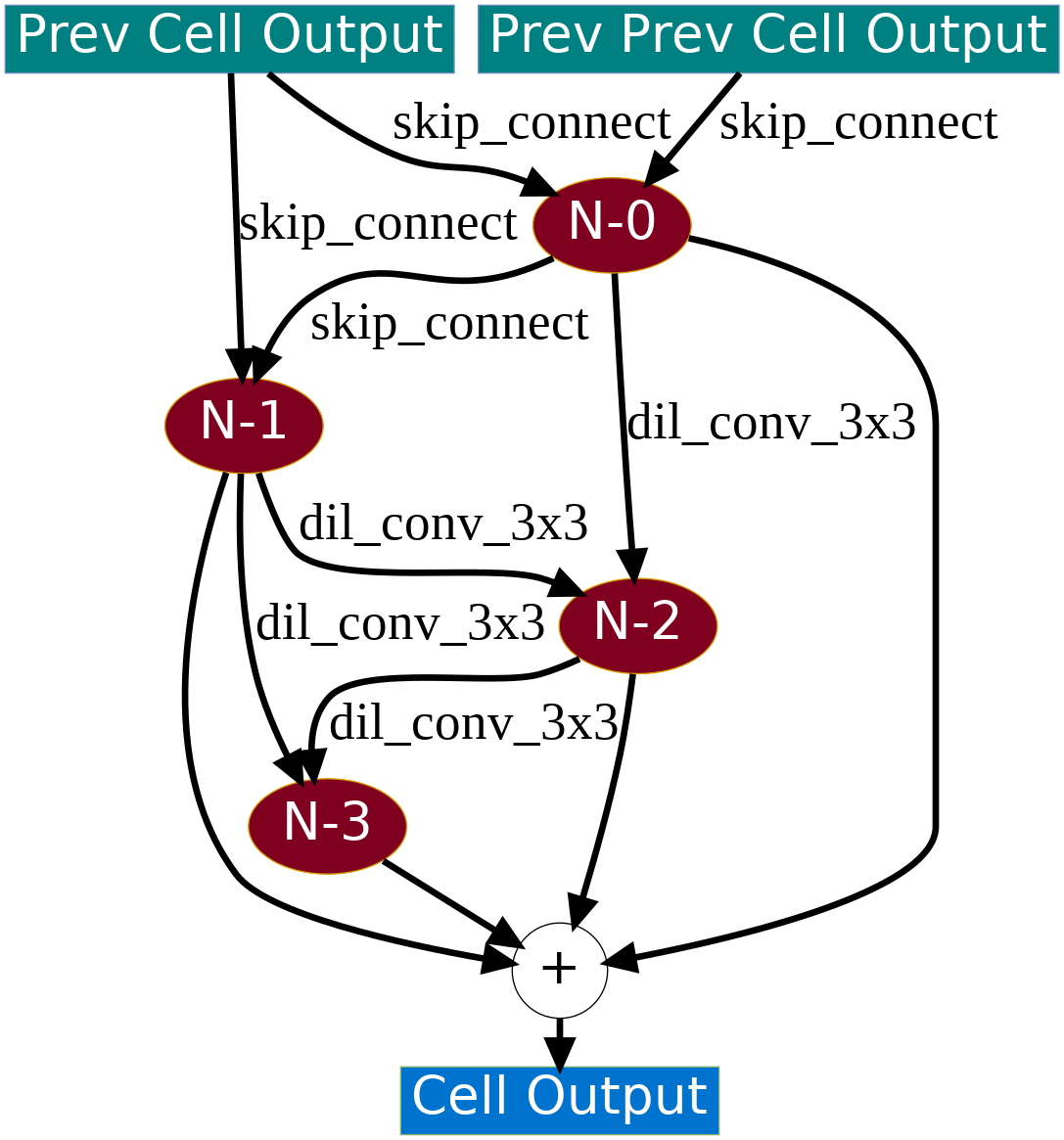}
  \hfill
  \caption{}
  \label{fig:7e}
\end{subfigure}
\begin{subfigure}{0.48\linewidth}
  \centering
  \includegraphics[width=1\textwidth, height=1.3in]{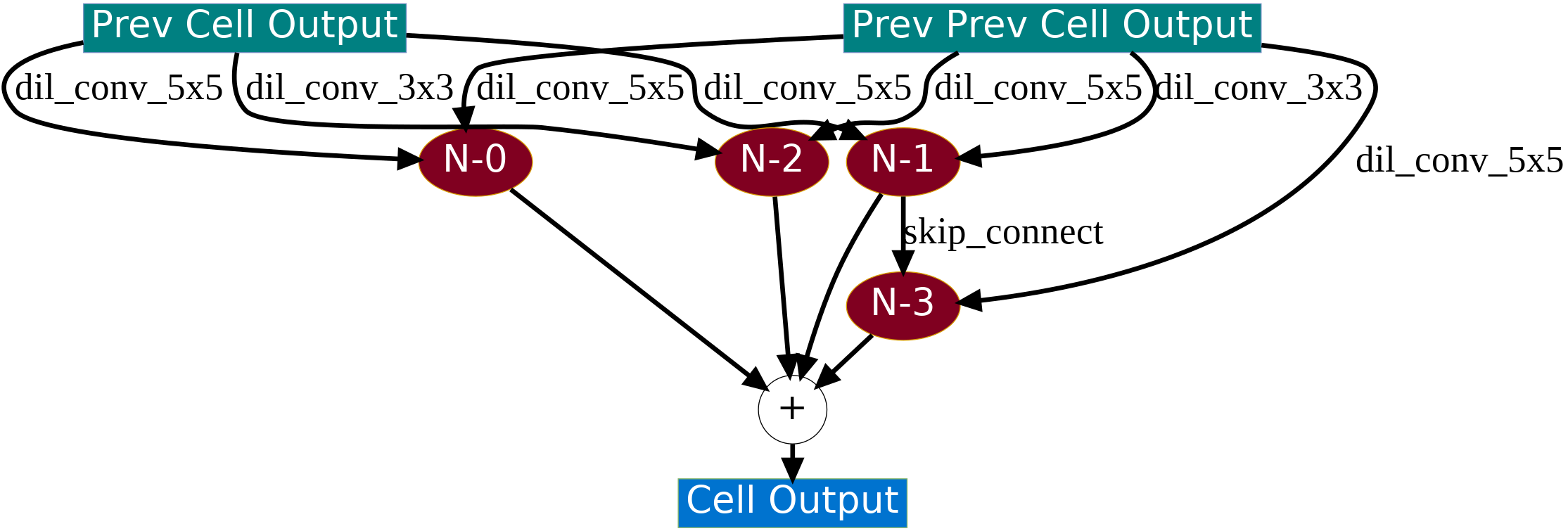}
  \hfill
  \caption{}
  \label{fig:7f}
\end{subfigure}

\caption{The final regular and reduction cell structures discovered by EmoNAS on (a,b) CASME-II and (c,d) SAMM.}
\label{fig:fig_micro}
\end{figure*}
\subsubsection{Datasets and Evaluation Settings} To evaluate the proposed EmoNAS, we conduct experiments over 7 macro-expression datasets CK+ \cite{lucey2010extended}, MUG \cite{aifanti2010mug}, ISED \cite{Happy2017TheIS}, DISFA \cite{mavadati2013disfa}, OULU-CASIA \cite{zhao2011facial}, FER2013 \cite{fer2013}, and RAF-DB~\cite{li2017reliable}. The sample facial expressions from these datasets {are} depicted in Figure~\ref{fig:dataset_samples}. We adopted a very strict experiment setting of subject/person independent evaluation. The models are tested over completely unseen subjects. This makes the achieved results very reliable. For FER2013, we follow the standard train-test scheme presented in the literature\cite{HappyFHOFH}.
The input image size of $120\times120$ is used for CK+, MUG, ISED and DISFA. For OULU-VIS-Strong and FER2013, we use the image size of $128\times128$ and $48\times 48$, respectively. The final cell structures discovered by EmoNAS for CK+ (6-exp), MUG (6-exp) are depicted in Figure \ref{fig:fig_ck_dis_mug}.\par

We also conduct experiments on 6 micro-expression datasets CASME-I \cite{yan2013casme}, CASME-II \cite{yan2014casme}, CAS(ME)\^2 \cite{qu2017cas}, SAMM \cite{davison2016samm}, SMIC~\cite{li2013spontaneous}, Composite (MEGC2019 challenge) dataset~\cite{see2019megc}. The video sequences in these datasets are preprocessed to create dynamic image \cite{bilen2016dynamic}. We use input size of $120\times 120$ for all 6 datasets. The final cell structures discovered by EmoNAS for CASME-I, CASME-II, CAS(ME)\^2, and SAMM are depicted in Figure~\ref{fig:fig_micro}. Similar to macro-expression, the models are evaluated on completely unseen subjects in leave-one-person-out manner.

\begin{table*}[]
\footnotesize
    \centering
    \caption{Comparative results of the proposed EmoNAS and existing state-of-the-art FER methods on MUG, ISED, DISFA dataset}
    \label{tab:tab_ised_dis_mug}
    \resizebox{\textwidth}{!}{
\begin{tabular}{l l c c c c c c }
        \toprule
        & & \multicolumn{2}{c}{\textbf{\textit{MUG}}}
       & \multicolumn{2}{c}{\textbf{\textit{ISED}}} & \multicolumn{2}{c}{\textbf{\textit{DISFA}}}\\
        \cmidrule(r){3-4}\cmidrule(r){5-6}\cmidrule(r){7-8}
        \textbf{\textit{Method}} &\textbf{\textit{Pub-Yr}}& {\textit{6 Exp}} & {\textit{7 Exp}} & {\textit{4 Exp}} & {\textit{5 Exp}}  & {\textit{6 Exp}} & {\textit{7 Exp}} \\
        \midrule
        RADAP \cite{Mandal2019RegionalAA}& IP-19  & 82.65 & 78.57 & 67.05 & N/A & 62.38 & 58.71 \\
        sLSP+LB \cite{rivera-2020} & TAFF-2- & N/A & N/A & 78.03& N/A&N/A&N/A\\
        VGG16 {$^{12}$} \cite{Simonyan2014VeryDC}& Arxiv-15 & 85.14 & 84.67 & 73.09 & 59.32 & 56.76 & 57.42 \\
        VGG19 {$^{12}$} \cite{Simonyan2014VeryDC}& Arxiv-15 & 85.22 & 85.12 & 70.24 & 69.20 & 59.98 & 53.66 \\
        ResNet50 {$^{12}$} \cite{He2015DeepRL}& CVPR-16 & 86.88 & 85.58 & 68.10 & 63.47 & 62.48 & 54.01 \\
        HiNet \cite{verma2019hinet}& LCS-19 & 87.80 & 87.20 & N/A & N/A & N/A & N/A \\
        DARTS {*} \cite{liu2018darts}& ICML-19 &94.39 & 93.70 & 76.87 & 47.29 & 66.55 & 49.98 \\
        P-DART {*}  \cite{chen2019progressive}& ICCV-19 &67.19 &91.16 &71.43 &53.50 &45.14 &33.58 \\
        \midrule
        \textbf{EmoNAS} & \textbf{Ours} &\textbf{95.60} & \textbf{96.55} & \textbf{79.93} & \textbf{65.88} & \textbf{69.18} & \textbf{61.2}\\
        \bottomrule
    \end{tabular}}
\end{table*}
\subsection{Macro-Expression Results Analysis}
\label{macro_results}
We compute the classification accuracy to evaluate the proposed EmoNAS on the macro-expression datasets CK+, MUG, ISED, DISFA, OULO-VIS-Strong, FER2013, and RAF-DB. The results on all the datasets are shown in Table \ref{tab:tab_ck+}, Table \ref{tab:tab_ised_dis_mug}, Table \ref{tab:tab_oulu}, Table~\ref{tab:tab_fer2013}, Table~\ref{tab:raf-db}. The results for VGG16, VGG19, ResNet50 and MobileNet were collected from~\cite{Mandal2019RegionalAA} and~\cite{verma2019hinet}.

\subsubsection{CK+} 
The proposed method is compared with existing deep learning methods~\cite{Xie2019FacialER,li2017reliable,Lopes2017FacialER,Yang2018IdentityAdaptiveFE,Liu2014FacialER,Khorrami2015DoDN,Cai2019IdentityFreeFE,verma2019hinet} in Table~\ref{tab:tab_ck+}. The architecture discovered by the proposed EmoNAS outperforms the existing state-of-the-art deep learning methods. The proposed NAS-based approach leads to the discovery of lightweight architectures as compared to the existing FER methods (refer Table~\ref{tab:tab_complex_comp}). From Table~\ref{tab:tab_complex_comp} we can observe that our model consists of lesser number of parameters ($\sim$0.55 M) in comparison to manually designed lightweight networks such as MobileNet ($\sim$3.2 M) and HiNet~\cite{verma2019hinet} ($\sim$1 M). In addition, it outperforms the existing methods in quantitative performance as well. EmoNAS obtains 5.73\% and 7.7\% higher recognition accuracy than HiNet in 6-class and 7-class FER, respectively. 
Similarly, it also outperforms the existing NAS based method DARTS~\cite{liu2018darts} (by 5.63\%, 1.29\%) and P-DARTS~\cite{pdarts} (by 11.2\%, 9.45\%) in 6- and 7-class problems, respectively. 
The results show that straightforward use of NAS algorithms (designed for generic classification problems)
is not sufficient to obtain good performance in FER problems. However, a well-designed NAS method, keeping the challenges in FER in mind, can achieve significantly better performance than the traditional CNN methods. from the results, we can observe that the Auto-FERNet is gaining more accuracy as compared to the proposed EmoNAS for CK+ dataset. However, Auto-FER requires approximately double computational power as compared to the proposed EmoNAS as tabulated in Table \ref{tab:tab_complex_comp}. {Also, recent vision transformer-based models \cite{aouayeb2021learning, kim2022facial} are achieving better performance on 7 expression class problems. Still, vision transformer-based models \cite{kim2022facial} need a huge computational power than compared to the proposed EmoNAS as reported in Table \ref{tab:tab_complex_comp}} 

\begin{table}[]
\footnotesize
\centering
    \caption{Comparative results of the proposed EmoNAS and existing state-of-the-art methods on OULU-VIS Strong dataset}
    \label{tab:tab_oulu}
    \begin{tabular}{l c c c c }
        \toprule
        & &\multicolumn{2}{c}{\textbf{\textit{OULU-VIS Strong}}}\\
        \cmidrule(r){3-4}
        \textbf{\textit{Method}} & \textit{\textbf{Pub-Yr}} & {\textit{6 Exp}} & {\textit{7 Exp}} \\
        \midrule
        RADAP \cite{Mandal2019RegionalAA}& IP-19 & 75.83 & 74.11  \\
        VGG16 {$^{12}$} \cite{Simonyan2014VeryDC}& Arxiv-15 & 73.40 & 70.70  \\
        VGG19 {$^{12}$} \cite{Simonyan2014VeryDC}&Arxiv-15 &71.4 & 70.5 \\
        ResNet5 {$^{12}$} \cite{He2015DeepRL}& CVPR-16 & 73.1 & 65.4  \\
        MobileNet {$^{7}$} \cite{Howard2017MobileNetsEC}& Arxiv-17 & 60.9 & 60.4  \\
        HiNet \cite{verma2019hinet}& LCS-19 & 70.3 & 72.0  \\
        \midrule
        \textbf{EmoNAS} & \textbf{Ours} &\textbf{93.54} & \textbf{90.71}\\
        \bottomrule
    \end{tabular}
\end{table}

\begin{table}[t!]
\footnotesize
\centering
    \caption{Comparative results of the proposed EmoNAS and existing state-of-the-art FER methods on FER2013 dataset}
    \label{tab:tab_fer2013}
    \medskip
 \begin{tabular}{l c c c c }
        \toprule
        \textbf{\textit{Method}} & \textit{\textbf{Pub-Yr}}& {\textbf{\textit{7 Exp (pr)}}} \\
        \midrule
        Bag of Words \cite{Ionescu2013LocalLT}& ICML-13 &67.4\\
        Mollahosseini \cite{Mollahosseini2016GoingDI}& WACV-16 &66.4 \\ 
        CNN-Ensemble \cite{Liu2016FacialER}&CW-16 &65.0 \\
        VGG+SVM \cite{Georgescu2018LocalLW}& IA-18 &66.3 \\
        GoogleNet \cite{Giannopoulos2018DeepLA}& AHIM-18 &65.2 \\
        Fa-Net \cite{Wang2019AFF}& Arxiv-19 &62.3 \\
        Auto-FERNet \cite{li2021auto}&TNSE-21 &\textbf{73.1} \\
        \midrule
        \textbf{EmoNAS} & \textbf{Ours} &\textbf{67.9}\\
        \bottomrule
    \end{tabular}
\end{table}

\begin{table*}[t!]
\footnotesize
\centering
\caption{Comparative results of the proposed EmoNAS and existing methods on RAF-DB dataset}
\label{tab:raf-db}
\medskip
\resizebox{\textwidth}{!}{
 \begin{tabular}{l c c c c c c c c c}
    \toprule
    \textbf{\textit{Method}}	 & \textit{\textbf{Pub-Yr}} &\textbf{\textit{Angry}} 	&\textbf{\textit{Disgust}} 	&\textbf{\textit{Fear}}	 &\textbf{\textit{Happy}} 	&\textbf{\textit{Sad}}	 &\textbf{\textit{Surprise}}	 &\textbf{\textit{Neutral}} 	&\textbf{\textit{Acc.}}\\
    \midrule
baseDCNN~\cite{li2017reliable}	&CVPR-17 &70.99	&52.50	&50.00	&92.91	&77.82	&79.64	&83.09	&82.66\\
Center Loss~\cite{li2017reliable} &CVPR-17	&68.52	&53.13	&54.05	&93.08	&78.45	&79.63	&83.24	&82.86\\
DLP-CNN~\cite{li2017reliable} &CVPR-17 &71.60	&52.15	&62.16	&92.83	&80.13	&81.16	&80.29	&82.74\\
VGG-FACE~\cite{fan2018multi} &ICANN-18	&82.19	&56.62	&55.41	&86.38	&79.52	&83.93	&71.18	&79.16\\
MRE-VGG~\cite{fan2018multi} &ICANN-18 	&83.95	&57.50	&60.81	&88.78	&79.92	&86.02	&80.15	&82.63\\
Kuo et al.~\cite{kuo2018compact} &CVPRW-18 &82.07	&44.59	&41.25	&81.01	&44.14	&90.12	&75.44	&72.21\\
FSN~\cite{zhao2018feature} &BMVC-18 &72.80	&46.90	&56.80	&90.50	&81.60	&81.80	&76.90	&81.14\\
PG-CNN~\cite{li2018occlusion} &TIP-19 &-	&-	&-	&-	&-	&-	&-	&83.27\\
SPWFA-SE~\cite{li2020facial} &TAFFC-20 &-	&-	&-	&-	&-	&-	&-	&86.31\\
DSAN-VGG~\cite{fan2020facial} &TAFFC-20	&82.71	&56.25	&58.11	&94.01	&83.89	&89.06	&80.00	&85.37\\
DSAN-RES~\cite{fan2020facial} &TAFFC-20	&82.10	&62.50	&55.41	&93.00	&84.94	&79.94	&83.97	&85.27\\
ViT-SE \cite{aouayeb2021learning}& Arxiv-21 &-&-&-&-&-&-&-&87.22\\
        Squeeze ViT \cite{kim2022facial}& Sensors-21 &-&-&-&-&-&-&-&\textbf{88.90}\\
        DAN \cite{wen2021distract}&Arxiv-21 &-&-&-&-&-&-&-&85.32\\
        DPOSTER \cite{zheng2022poster}& Arxiv-21 &-&-&-&-&-&-&-&86.03\\
\textbf{EmoNAS} & \textbf{Ours}	&\textbf{83.78}	&\textbf{86.02}	&\textbf{70.62}	&\textbf{77.77}	&\textbf{83.05}	&\textbf{94.85}	&\textbf{82.37}	&\textbf{87.28}\\
 
        \bottomrule
    \end{tabular}}
\end{table*}

\subsubsection{MUG, ISED and DISFA} 
We further evaluate the quantitative results for ISED, DISFA, and MUG in Table \ref{tab:tab_ised_dis_mug}.
The MUG consists of posed facial expressions whereas, ISED and DISFA consist of spontaneous expressions. for MUG, the proposed EmoNAS beats the other NAS approaches DARTS~\cite{liu2018darts} (by 1.21\%, 2.85\%) and P-DARTS~\cite{pdarts} (by 28.41\%, 5.39\%) in 6- and 7-class FER, respectively. In ISED, our method substantially improves upon the DARTS~\cite{liu2018darts} (by 2.76\% and 18.59\%) and P-DARTS~\cite{pdarts}(by 8.5\% and 12.38\%) in 4- and 5-class problems, respectively. Furthermore, in DISFA, the proposed EmoNAS outperforms DARTS~\cite{liu2018darts} (by margins of 2.63\% and 11.22\%) and P-DARTS~\cite{pdarts} (by margins of 24.04\% and 27.62\%) in 6- and 7-class problems, respectively. In comparison to the traditional designed CNN networks, the proposed NAS based models show improvement in both the quantitative performance (Table \ref{tab:tab_ised_dis_mug}) and computational complexity (Table~\ref{tab:tab_complex_comp}). The results also show that the proposed method is effective for both posed and spontaneous FER problems.\par

\subsubsection{OULU-VIS-Strong and FER 2013} The quantitative results in OULU-VIS-Strong are shown in Table~\ref{tab:tab_oulu}. From Table \ref{tab:tab_oulu}, it is evident that the proposed models derived from EmoNAS obtain superior performance over existing approaches. Our model achieves 23.24\%, 18.71\% improvement over HiNet in 6- and 7-class FER, respectively. We also evaluate our model over the private test set of FER2013 in the 7-class setting in Table \ref{tab:tab_fer2013}. The FER 2013 is a challenging benchmark FER dataset. The previous methods~\cite{Mollahosseini2016GoingDI,Liu2016FacialER,Wang2019AFF,Giannopoulos2018DeepLA,Georgescu2018LocalLW,Ionescu2013LocalLT} have used the traditional CNN networks to improve the performance. Our NAS-based approach clearly outperforms the existing state-of-the-art approaches~\cite{Mollahosseini2016GoingDI,Liu2016FacialER,Wang2019AFF,Giannopoulos2018DeepLA,Georgescu2018LocalLW,Ionescu2013LocalLT}. This further shows the robustness of the proposed EmoNAS to a variety of FER datasets collected from a diverse set of scenarios.\par

\subsubsection{RAF-DB} 
 We also conduct experiments on the in-the-wild RAF-DB dataset.
 We search and train the EmoNAS with the seven expression labels provided in the RAF-DB. Table~\ref{tab:raf-db} shows the class-wise accuracy and the overall accuracy obtained by EmoNAS for RAF-DB. The class-wise accuracy for anger, disgust, fear, happy, sad, surprise, and neutral (83.78\%, 86.02\%, 70.62\%, 77.77\%, 83.05\%, 94.85\%, 82.37\%) show that each specific expression is correctly recognized. We also compare our results with the existing state-of-the-art approaches. The proposed EmoNAS (87.28\% overall accuracy) outperforms the {all most all other} methods in the literature. More specifically, it obtains 2.01\%, 1.91\%, 0.97\%, 4.01\% improvement over the recent works published in TIP-19 (PG-CNN~\cite{li2018occlusion}), TAFFC-20 (SPWFA-SE~\cite{li2020facial}), TAFFC-20 (DSAN-VGG~\cite{fan2020facial}), TAFFC-20 (DSAN-RES~\cite{fan2020facial}), respectively. Furthermore, the results reveal that the EmoNAS attains the highest accuracy when identifying the category of disgust, fear, and surprise. It also maintains good accuracy in the remaining categories, leading to overall accuracy, better than the existing state-of-the-art. Therefore, the features learned and chosen by EmoNAS should contain more discriminative information for FER. This illustrates the effectiveness and superiority of the NAS-based search for the most robust architecture over the traditional CNN models.

\begin{table}[]
\footnotesize
    \centering
    \resizebox{\columnwidth}{!}{
    \begin{threeparttable}
    \begin{tabular}{l c c c c }
        \toprule
        \textbf{Method} &\textbf{Pub-Yr}& \textbf{Task} & \textbf{CASME-I} & \textbf{CASME-II}\\
        \midrule
        MDMO~\cite{liu2015main}& TAFFC-15 &[P, N, S, O] & 56.29 & 51.69\\
        SparseMDMO~\cite{sparseMDMO}& TAFFC-18 & [P, N, S, O] & 74.83 & 66.95\\
        CNN-LSTM~\cite{CNN-LSTM}& Neuro-18 & [H, S, D, R, O] & 60.98 & N/A\\
        3D-Flow~\cite{Li3D-Flow}& PAA-19 & [H, S, D, R, T] & 55.44 & 59.11\\
        Spatio-Temp~\cite{Kimspatiotemporal}& TAFFC-17 & [H, S, D, R, O] & N/A & 60.98 \\
        FuseNet  {*}\cite{ReddyFuse}& IJCNN-18 & [P, N, S, O] & 54.84 & 45.11 \\
        3D-ResNet34 {*} \cite{He2015DeepRL}& CVPR-16 & [P, N, S, O] & 55.21 & 35.21\\
        3D-ResNet50 {*} \cite{He2015DeepRL}& CVPR-16 & [P, N, S, O] & 56.26 & 35.78\\
        OrigiNet \cite{monu-ijcnn} & IJCNN-20 & [P, N, S, O] & 66.30 & 61.58 \\
        Aff.Net \cite{monu-multimedia} & IEEEMM-21 & [P, N, S, O] & 66.99 & 61.58\\
        DARTS {*} \cite{liu2018darts}& ICLR-19 & [P, N, S, O] & 73.72 & 59.63\\
        P-DART {*} \cite{chen2019progressive}& ICCV-19 & [P, N, S, O] & 57.03 & 64.38 \\
       AutoMER~\cite{verma2021automer}&TNNLS-21 & [P, N, S, O] &{77.58} & \textbf{74.15} \\
        \midrule
         {ADL\_Class} &  {Ablation} &  {[P, N, S, O]} &  {76.04} &  {\textbf{\textit{75.35}}} \\
        \textbf{EmoNAS}& \textbf{Ours} & \textbf{[P, N, S, O]} & \textbf{80.00} &\textbf{68.42}\\
        \bottomrule
    \end{tabular}\begin{tablenotes}[para,flushleft]
   \textit{ {H: Happy, S: Surprise, D: Disgust, R: Regression, O: Others, P: Positive, N: Negative.} 
   }
  \end{tablenotes}
  \end{threeparttable}
    }
    \caption{Comparative results of the proposed EmoNAS and existing approaches on CASME-I and CASME-II}
    \label{tab:tab_micro1}
\end{table}

\begin{table}[]
\footnotesize
\centering
 \resizebox{\columnwidth}{!}{
    \begin{tabular}{l c c c c }
        \toprule
        \textbf{Method} & \textbf{Pub-Yr} &\textbf{Task} & \textbf{CAS(ME)\^2} & \textbf{SAMM}\\
        \midrule
        OrigiNet \cite{monu-ijcnn}& IJCNN-20 & 3/4 EXP & 52.85 & 47.46\\
        Aff.Net~\cite{monu-multimedia} & IEEEMM-21 & 3/4 EXP & 54.56 & 34.89\\
        FuseNet {*}~\cite{ReddyFuse} &IJCNN-19 & 3/4 EXP & 50.77 & 46.72 \\
        3D-ResNet34 {*}~\cite{He2015DeepRL}&CVPR-16 & 3/4 EXP & 44.59 & 47.13\\
        3D-ResNet50 {*}~\cite{He2015DeepRL}&CVPR-16 & 3/4 EXP & 43.09 & 49.89\\
        DARTS {*}~\cite{liu2018darts}&ICLR-19 & 3/4 EXP & 58.20 & 76.98\\
        P-DART {*}~\cite{chen2019progressive}&ICCV-19 & 3/4 EXP &58.19 & 72.88 \\
         AutoMER~\cite{verma2021automer}&TNNLS-21 &3/4 EXP &\textbf{78.08} & {72.45} \\
        \midrule
         {ADL\_Class} &  {Ablation}&  {3/4 EXP}&  {70.78} &  {{75.11}} \\
        \textbf{EmoNAS}&\textbf{Ours} & \textbf{3/4 EXP} &
        \textbf{71.15} &\textbf{77.04}\\
        \bottomrule
    \end{tabular}}
    \caption{Comparative results of the proposed EmoNAS and existing approaches on CASME\^2 and SAMM.}
    \label{tab:tab_micro2}
\end{table}

\subsection{Micro-Expression Results Analysis}
\label{micro_results}
Usually, the macro expression recognition algorithms rely on static images for analysis. Whereas, the micro-expression (MEs) recognition algorithms require a video sequence as input. 
For micro-expression, we use dynamic imaging to generate a single instance of an image micro-expressions. 
DI ensures uniform search and training architecture for both macro and micro-expression recognition problems. We use the generated dynamic images to search and train the NAS models for micro-expression.\par

\subsubsection{CASME-I and CASME-II}
For micro-expression, the effectiveness is examined by testing in terms of average classification accuracy. CASME-I dataset consists of 195 video clips of 35 subjects of Chinese ethnicity. We cataloged the micro-expressions into 4 groups: positive-12, negative-50, surprise-21 and others-106 as given in \cite{verma2019learnet}. Similarly, CASME-II videos are grouped into positive-12, negative-50, surprise-21, and others-106. Table \ref{tab:tab_micro1} shows the classification accuracy results for CASME-I and CASME-II. We compare our results with both handcrafted and deep learning methods in the literature. A recently published work in~\cite{xia2020revealing} re-evaluated several previous methods over the leave-one-subject-out (LOSO) protocol for a fair comparison. In our experiments, we follow a similar setup and compare our proposed method with the comparative results as presented in~\cite{xia2020revealing}. The proposed EmoNAS significantly outperforms the existing methods in recognition accuracy. It also improves upon the existing NAS methods DARTS~\cite{liu2018darts} (by 6.28\%, 8.79\%) and P-DARTS~\cite{pdarts} (by 22.97\%, 4.04\%) in CASME-I and CASME-II, respectively. 
\begin{table}[t!]
\footnotesize
\centering
\caption{Recognition accuracy and unweighted average recall (UAR) comparison on SMIC and Composite dataset (MEGC2019 Challenge)}
\label{tab_smic_composite}
\resizebox{\columnwidth}{!}{%
\begin{tabular}{l l c c c}
\hline \noalign{\smallskip}
 \multicolumn{1}{l}{\multirow{2}{*}{\textbf{\textit{Method}}}} & \multicolumn{1}{l}{\multirow{2}{*}{\textbf{\textit{Pub-Year}}}} &\multicolumn{2}{c}{\textbf{\textit{SMIC}}}                               & \multicolumn{1}{c}{\textbf{\textit{COMPOSITE}}} \\ \cline{3-5}  \noalign{\smallskip}
\multicolumn{1}{l}{}               &\multicolumn{1}{l}{}         & \multicolumn{1}{c}{\textbf{\textit{Acc.}}} & \multicolumn{1}{c}{\textbf{\textit{UAR}}} & \multicolumn{1}{c}{\textbf{\textit{UAR}}}    \\ \hline \noalign{\smallskip}
 LBP-TOP \cite{Zhao}  &A-Tran-07                 & N/A      & 52.80        & 57.85      \\
  MDMO~\cite{liu2015main} &TAFFC-15                 & 64.00      & 56.91       & N/A       \\
 OffApexNet \cite{gan2019off}  & SP-IC-19          & 67.68      & N/A         & N/A       \\
CNN-LSTM \cite{wang2018micro} &   Ar-18          & N/A        & 41.25       & 39.24     \\
TSCNN  \cite{song2019recognizing}& Access-19              & N/A        & 54.12       & 59.48     \\
 STSTNet \cite{liong2019shallow} &FG-19             & N/A        & 59.95       & 67.24     \\
  NMER \cite{liu2019neural}  & FG-19              & N/A        & 55.55       & 59.36     \\
 DualInc \cite{zhou2019dual}  & FG-19             & N/A        & 61.49       & 68.58     \\
 CapsuleNet \cite{van2019capsulenet} &FG-19            & N/A        & 58.77       & 65.06     \\
 DCN-DB \cite{xia2019cross}    &ICBEA-19           & N/A        & 59.79       & 60.37     \\
 DSSN \cite{khor2019dual}   &  ICIP-19             & 63.41      & N/A         & N/A       \\
 RCN-W  \cite{xia2020revealing}& TIP-20               & N/A        & 66.00       & 71.00    \\
  LGCconD \cite{li2020joint} & TIP-20             & 63.41      & N/A         & N/A       \\
 AutoMER \cite{verma2021automer} &TNNLS             & \textbf{81.20}      & N/A         & N/A       \\
  \hline  \noalign{\smallskip}
    {ADL\_Class} &  {Ablation}&  {71.39}&  {58.54} &  {\textbf{\textit{78.76}}} \\
 \textbf{EmoNAS}  &\textbf{ Ours}          & \textbf{77.09}      & \textbf{69.03}       &\textbf{ {73.22} }\\\hline       
\end{tabular}%
}
\end{table}
\subsubsection{CAS(ME)\^2 and SAMM}
CAS(ME)\^2 consists of 339 sequences with three emotion classes: anger-101, happy-149, and disgust-88. SAMM includes 159 spontaneous micro-expressions of 29 subjects, recorded at 200 fps. Similar to CASME-I and CASME-II, we combine similar emotion classes to augment 4 groups: positive-27, negative-80, surprise-15, and others-37. Table \ref{tab:tab_micro2} shows the classification accuracy results for CAS(ME)\^2 and SAMM. Our method outperforms the existing methods. More specifically, it outperforms the FuseNet~\cite{ReddyFuse} by 20.38\% and 30.32\% in CAS(ME)\^2 and SAMM, respectively. It also improves upon DARTS~\cite{liu2018darts} (by a margin of 12.95\% and 0.06\%) and PDARTS~\cite{pdarts} (by a margin of 12.96\% and 4.16\%) in CAS(ME)\^2 and SAMM, respectively. 
\begin{table}[t!]
\footnotesize
\centering
    \caption{EmoNAS ablation analysis on CK+ with 6 expressions}
    \label{tab:ck+_abl}
     \begin{tabular}{l c c c }
        \toprule
       \multirow{2}{*}{\textbf{\textit{\#Cells}}} & \multicolumn{3}{c}{\textit{\textbf{\#Nodes}}}\\
        \cmidrule(r){2-4}
        & 6 & 7 & 8\\
        \midrule  
        5 & 95.67 & \textbf{97.13} & {95.96} \\
         8 & 84.93  &  85.23 & 83.88  \\
        10 & {96.15} & 96.04 & 95.84 \\
        15 & 94.78 & 96.11 & 93.42\\
        \bottomrule
    \end{tabular}
 \end{table}
\subsubsection{SMIC and Composite Dataset}
We also conduct experiments on SMIC and the Composite dataset~\cite{see2019megc}. SMIC consists of 164 emotion sequences from 16 subjects. The emotions are grouped into positive-51, negative-70, and surprise-43. The Composite dataset is collected by combining three emotion classes (negative, positive, and surprise) of three datasets CASME-II, SAMM, and SMIC as given in MEGC-2019~\cite{see2019megc} challenge.
The quantitative results for SMIC and Composite datasets are given in Table~\ref{tab_smic_composite}. In SMIC, the proposed EmoNAS obtains superior recognition accuracy and unweighted average recall (UAR) over the existing methods. It outperforms the most recent method LGCconD~\cite{li2020joint} (TIP-20) in accuracy by a margin of 13.68\%. Similarly, it outperforms RCN-W~\cite{xia2020revealing} (TIP-20) in UAR by 3.03\% margin. In the Composite dataset, our EmoNAS achieves 2.22\% higher UAR as compared to the current state-of-the-art method RCN-W~\cite{xia2020revealing} (TIP-20).

 \begin{table}[t!]
\footnotesize
\centering
    \caption{EmoNAS ablation analysis on DISFA with 6 expressions}
    \label{tab:disfa_abl}
    \begin{tabular}{l c c c }
        \toprule
       \multirow{2}{*}{\textbf{\textit{\#Cells}}} & \multicolumn{3}{c}{\textit{\textbf{\#Nodes}}}\\
        \cmidrule(r){2-4}
        & 6 & 7 & 8\\
        \midrule
        5 & 63.89 & 69.18 & 72.57 \\
         8 &  {68.84} & {73.28}  &  71.54 \\
        10 &  64.63 & 69.89 &\textbf{73.37} \\
        15 & 65.68 & 67.17 & 71.03 \\
        \bottomrule
    \end{tabular}
\end{table}
\subsection{Ablation Study}
\label{ablation}
We study the effect of multiple components of EmoNAS through ablation analysis to understand its behavior for the task of FER. We conduct experiments on two macro-expression datasets CK+, DISFA, and two micro-expression datasets. The ablation results are shown in Table \ref{tab:ck+_abl}, Table \ref{tab:disfa_abl}, Table \ref{tab:casme1_abl}, and Table \ref{tab:casme_sq_abl}.  { We change the number of nodes within each cell to 6, 7, and 8 nodes. We stack these cells with 5, 8, 10, and 15 repetitions to obtain 12 different CNN structures. Thus, for each dataset, we conduct 12 ablation experiments.}\par
{ {From Table \ref{tab:ck+_abl} and Table \ref{tab:disfa_abl} we can see that increasing the number nodes to 8 or decreasing to 6 resulted in marginal decrease or increase in the accuracy for CK+ and DISFA. Similarly, increasing the number of cells to 10/15 doesn't improve the performance substantially in both datasets. Moreover, a higher number of cells would increase the network complexity. Therefore, in most cases, we select 5 cells with each having 7 nodes to maintain a good trade-off between accuracy and model efficiency. The micro-expression ablations results are given in Table~\ref{tab:casme1_abl} and Table~\ref{tab:casme_sq_abl}. The best performance is achieved with 5 cells and 7 nodes. This again shows that the combination of 5 cells and 7 nodes are the most suitable parameters for FER datasets. These results also prove the literature \cite{verma2019learnet, pasupa2016comparison, srivastava2015training} deep networks fail to achieve adequate performance over small-sized datasets.} In addition, we conducted an ablation study with the architecture search of the autodeeplab \cite{liu2019auto} for the classification (ADL\_Class). The experiments are conducted over the five micro expression datasets CASME-I, CASME-II, CAS(ME)\^2, SMIC, SAMM along with Composite dataset and the results are included in Table \ref{tab:tab_micro1}, \ref{tab:tab_micro2} and \ref{tab_smic_composite}. }
\begin{table}[t!]
\footnotesize
\centering
    \caption{EmoNAS ablation analysis on CASME-I}
    \label{tab:casme1_abl}
     \begin{tabular}{l c c c }
        \toprule
       \multirow{2}{*}{\textbf{\textit{\#Cells}}} & \multicolumn{3}{c}{\textit{\textbf{\#Nodes}}}\\
        \cmidrule(r){2-4}
        & 6 & 7 & 8\\
        \midrule  
        5 & 69.50 & \textbf{80.00} & 70.65 \\
         8 &  {79.29} &  74.31 & 73.46  \\
        10 & 73.37 & {77.19} & {76.17} \\
        15 & 75.74 & 72.11 & 74.52\\
        \bottomrule
    \end{tabular}
\end{table}
 
\begin{table}[t!]
\footnotesize
\centering
    \caption{EmoNAS ablation analysis on CAS(ME)\^2}
    \label{tab:casme_sq_abl}
    \begin{tabular}{l c c c }
        \toprule
        \multirow{2}{*}{\textbf{\textit{\#Cells}}}& \multicolumn{3}{c}{\textit{\textbf{\#Nodes}}}\\
        \cmidrule(r){2-4}
        & 6 & 7 & 8\\
        \midrule
        5 & 67.90 & 71.15 & 69.30 \\
         8 & {73.25}  & \textbf{74.95}  & {69.8}2  \\
        10 & 70.44 & 67.57 & 62.16 \\
        15 & 70.63 & 65.56 & 60.56 \\
        \bottomrule
    \end{tabular}
\end{table}

\begin{table}[]
\footnotesize
    \centering
    \caption{Memory, speed, and complexity analysis of EmoNAS models discovered on macro: CK$+$, MUG, ISED, DISFA, OULU-VIS-Strong, FER2013 and micro: CASME-I, CASME-II, CAS(ME)\^2, SAMM datasets.}
    \label{tab:tab_params}
    
    \begin{tabular}{l c c c c }
        \toprule
        \textbf{Dataset} & \textbf{Input Size} & \textbf{Size(mb)} & \textbf{Speed(fps)} & \textbf{Params(M)} \\
        \midrule
        CK$+$ & $120\times120$ & 15.45 & 60.30 & 0.90  \\
        MUG & $120\times120$ & 13.75 & 17.15 & 0.49  \\
        ISED & $120\times120$ & 13.93 & 56.25 & 0.51  \\
        DISFA & $120\times120$ & 14.15 & 30.90 & 0.58  \\
        OULU-VIS-Strong & $128\times128$ & 14.00 & 54.50 & 0.59 \\
        FER2013 & $48\times48$ & 7.43 & 200.80 & 1.36  \\
        CASME-I & $128\times128$ & 14.63 & 42.45 & 0.70  \\
        CASME-II & $180\times180$ & 16.43 & 32.63 & 0.82   \\
        CAS(ME)\^2 & $180\times180$ & 14.94 & 31.68 & 0.53  \\
        SAMM & $120\times120$ & 13.64 & 40.56 & 0.6  \\
        \bottomrule
    \end{tabular}
    
\end{table}

\begin{table}[t]
\footnotesize
    \centering
    \caption{Comparative analysis of the complexity of the EmoNAS with existing methods.}
    \label{tab:tab_complex_comp}
    \resizebox{\columnwidth}{!}{
    \begin{tabular}{l c c c c c}
    \toprule
Method	 & Pub-Year &Task &Size (mb)	 &Speed (fps)	 &Params (M) \\
\midrule
OrigiNet~\cite{monu-ijcnn}	 &IJCNN-20 & MER &14.3 &$\sim$0.22  &1.8\\
AffectiveNet~\cite{monu-multimedia}  &Mult-20 &MER &8.3	 &0.12	 &2.2 \\
MobileNet~\cite{verma2019hinet} 	 &LCS-19 &MER &25.3	 &0.08	 &4.2\\
DARTS~\cite{liu2018darts}  &ICLR-19 &MER &$\sim$16	 &$\sim$35	 &$\sim$0.9\\
P-DARTS~\cite{pdarts}  &ICCV-19 &MER &52	 &$\sim$5	 &$\sim$3.6\\

\textbf{EmoNAS} 	 &\textbf{Ours} &\textbf{MER} &\textbf{$\sim$14}	 &\textbf{$\sim$35}	 &\textbf{$\sim$0.60}\\ \midrule
MobileNet~\cite{verma2019hinet} 	 &LCS-19 &\textbf{FER} &23.8 &NA	&3.2\\
HiNet~\cite{verma2019hinet}  &LCS-19 &FER &5.3	 &NA	 &1.0\\
DARTS~\cite{liu2018darts}  &ICLR-19 &FER &17	 &$\sim$35	 &$\sim$1.0\\
P-DARTS~\cite{pdarts}  &ICCV-19 &FER &51	 &$\sim$5	 &$\sim$3.5\\
DARTS~\cite{liu2018darts}  &ICLR-19 &FER &$\sim$17	 &$\sim$35	 &$\sim$1.0\\
Auto-FERNet~\cite{li2021auto}  &TNSE-21 &FER &N/A	 &N/A	 &$\sim$2.1\\
ViT~\cite{kim2022facial}  &Sensors-22 &FER &N/A	 &N/A	 &$\sim$86.86\\
Squeeze ViT~\cite{kim2022facial}  &Sensors-22 &FER &N/A	 &N/A	 &$\sim$11.96\\
\textbf{EmoNAS} &\textbf{Ours} &\textbf{FER} &\textbf{$\sim$14}	 &\textbf{$\sim$45}	 &\textbf{$\sim$0.55}\\
\bottomrule
    \end{tabular}}
\end{table}

\subsection{Speed, Memory and Complexity analysis}
The models generated through EmoNAS search are very lightweight and fast. Table~\ref{tab:tab_params} shows the memory, speed, and computational complexity of the models generated by EmoNAS in different macro and micro-expression datasets. The proposed models require trainable parameters of approximately $\approx$ 0.5 Million (M) except for FER2013 and CK+. This is substantially lower than MobileNet (3.2M), VGG16 (138M), VGG19 (144M) ResNet50 (31M). It is even lower than HiNet (1M) with much better accuracy over the same. Also, our models use less memory space (~13-15MB) as compared to MobileNet (23.8 MB), VGG16 (500.3 MB), VGG19 (520.4 MB), and ResNet (88.4 MB). Hence, it can be useful for deployment in embedded devices. The speed over GPU shows that the proposed models are remarkably fast (up to 60 fps), which makes them suitable for real-time applications.\par
We further compare our work with the existing methods for both MER and FER in Table~\ref{tab:tab_complex_comp}. The EmoNAS models require the lowest number of operations for inference with $\sim$0.55-$\sim$0.60 million parameters in the model. Our model size is $\sim$38 MB less than the existing NAS method (P-DARTS). Similarly, the total number of parameters for EmoNAS is 6$\times$ less than the P-DARTS. The inference speed in MER and FER is $\sim$35 and $\sim$45, respectively. {Moreover, our proposed EmoNAS is also more efficient as compared to recent vision transformer based FER models as shown in Table ~\ref{tab:tab_complex_comp}. 
Particularly, proposed EmoNAS is 155$\times$ and 20$\times$  less than the ViT and Squeeze viT \cite{kim2022facial} FER models, respectively.} Thus, the optimized networks discovered by EmoNAS are suitable for embedded devices used in real-time applications.

\section{Conclusions}
To the best of our knowledge, this paper presents the first attempt at a NAS-based approach for the task of FER in both macro and micro-level facial expressions. We proposed EmoNAS which is based on the optimization techniques presented in DARTS to expedite the searching process. The design decisions for EmoNAS are the result of careful analysis of the FER domain-specific challenges. The same is validated by its superior performance over DART and P-DART methods. The architecture search also led to the formulation of shallower and lightweight CNNs. The resulting models achieve better performance compared to the existing state-of-the-art FER methods in 13 benchmark (7 FER and 6 MER) datasets. The searched models obtain higher accuracy with a fraction of the computational cost and very high inference speed. In the future, shared parameters-based schemes can be investigated in the search process to discover even better performance-aware architectures. Furthermore, the number of skip connections can be optimized separately for a better trade-off between complexity and performance. The decision of selecting the number of cells in evaluation could also be included as a parameter for optimization while searching. The proposed framework is designed to work with single image input. This requires pre-processing of the video data into a single instance feature map for MER analysis. In the future, we plan to design a framework to handle the image and raw video data input for both macro- and micro-expression problems.

\bibliography{template}

\begin{thebibliography}{97}
\expandafter\ifx\csname natexlab\endcsname\relax\def\natexlab#1{#1}\fi
\providecommand{\url}[1]{\texttt{#1}}
\providecommand{\href}[2]{#2}
\providecommand{\path}[1]{#1}
\providecommand{\DOIprefix}{doi:}
\providecommand{\ArXivprefix}{arXiv:}
\providecommand{\URLprefix}{URL: }
\providecommand{\Pubmedprefix}{pmid:}
\providecommand{\doi}[1]{\href{http://dx.doi.org/#1}{\path{#1}}}
\providecommand{\Pubmed}[1]{\href{pmid:#1}{\path{#1}}}
\providecommand{\bibinfo}[2]{#2}
\ifx\xfnm\relax \def\xfnm[#1]{\unskip,\space#1}\fi
\bibitem[{Aifanti et~al.(2010)Aifanti, Papachristou \&
  Delopoulos}]{aifanti2010mug}
\bibinfo{author}{Aifanti, N.}, \bibinfo{author}{Papachristou, C.}, \&
  \bibinfo{author}{Delopoulos, A.} (\bibinfo{year}{2010}).
\newblock \bibinfo{title}{The mug facial expression database}.
\newblock In {\it \bibinfo{booktitle}{11th International Workshop on Image
  Analysis for Multimedia Interactive Services WIAMIS 10}\/} (pp.
  \bibinfo{pages}{1--4}).
\newblock \bibinfo{organization}{IEEE}.
\bibitem[{Ambadar et~al.(2005)Ambadar, Schooler \&
  Cohn}]{ambadar2005deciphering}
\bibinfo{author}{Ambadar, Z.}, \bibinfo{author}{Schooler, J.~W.}, \&
  \bibinfo{author}{Cohn, J.~F.} (\bibinfo{year}{2005}).
\newblock \bibinfo{title}{Deciphering the enigmatic face: The importance of
  facial dynamics in interpreting subtle facial expressions}.
\newblock {\it \bibinfo{journal}{Psychological science}\/},  {\it
  \bibinfo{volume}{16}\/}, \bibinfo{pages}{403--410}.
\bibitem[{Aouayeb et~al.(2021)Aouayeb, Hamidouche, Soladie, Kpalma \&
  Seguier}]{aouayeb2021learning}
\bibinfo{author}{Aouayeb, M.}, \bibinfo{author}{Hamidouche, W.},
  \bibinfo{author}{Soladie, C.}, \bibinfo{author}{Kpalma, K.}, \&
  \bibinfo{author}{Seguier, R.} (\bibinfo{year}{2021}).
\newblock \bibinfo{title}{Learning vision transformer with squeeze and
  excitation for facial expression recognition}.
\newblock {\it \bibinfo{journal}{arXiv preprint arXiv:2107.03107}\/}, .
\bibitem[{Baker et~al.(2017)Baker, Gupta, Raskar \&
  Naik}]{baker2017accelerating}
\bibinfo{author}{Baker, B.}, \bibinfo{author}{Gupta, O.},
  \bibinfo{author}{Raskar, R.}, \& \bibinfo{author}{Naik, N.}
  (\bibinfo{year}{2017}).
\newblock \bibinfo{title}{Accelerating neural architecture search using
  performance prediction}.
\newblock {\it \bibinfo{journal}{arXiv preprint arXiv:1705.10823}\/}, .
\bibitem[{Bender et~al.(2018)Bender, Kindermans, Zoph, Vasudevan \&
  Le}]{Bender2018UnderstandingAS}
\bibinfo{author}{Bender, G.}, \bibinfo{author}{Kindermans, P.-J.},
  \bibinfo{author}{Zoph, B.}, \bibinfo{author}{Vasudevan, V.}, \&
  \bibinfo{author}{Le, Q.~V.} (\bibinfo{year}{2018}).
\newblock \bibinfo{title}{Understanding and simplifying one-shot architecture
  search}.
\newblock In {\it \bibinfo{booktitle}{ICML}\/}.
\bibitem[{Bilen et~al.(2016)Bilen, Fernando, Gavves, Vedaldi \&
  Gould}]{bilen2016dynamic}
\bibinfo{author}{Bilen, H.}, \bibinfo{author}{Fernando, B.},
  \bibinfo{author}{Gavves, E.}, \bibinfo{author}{Vedaldi, A.}, \&
  \bibinfo{author}{Gould, S.} (\bibinfo{year}{2016}).
\newblock \bibinfo{title}{Dynamic image networks for action recognition}.
\newblock In {\it \bibinfo{booktitle}{Proceedings of the IEEE Conference on
  Computer Vision and Pattern Recognition}\/} (pp.
  \bibinfo{pages}{3034--3042}).
\bibitem[{Cai et~al.(2019)Cai, Meng, Khan, Li, O'Reilly \&
  Tong}]{Cai2019IdentityFreeFE}
\bibinfo{author}{Cai, J.}, \bibinfo{author}{Meng, Z.}, \bibinfo{author}{Khan,
  A.-S.}, \bibinfo{author}{Li, Z.}, \bibinfo{author}{O'Reilly, J.}, \&
  \bibinfo{author}{Tong, Y.} (\bibinfo{year}{2019}).
\newblock \bibinfo{title}{Identity-free facial expression recognition using
  conditional generative adversarial network}.
\newblock {\it \bibinfo{journal}{ArXiv}\/},  {\it
  \bibinfo{volume}{abs/1903.08051}\/}.
\bibitem[{Chen et~al.(2019{\natexlab{a}})Chen, Xie, Wu \&
  Tian}]{chen2019progressive}
\bibinfo{author}{Chen, X.}, \bibinfo{author}{Xie, L.}, \bibinfo{author}{Wu,
  J.}, \& \bibinfo{author}{Tian, Q.} (\bibinfo{year}{2019}{\natexlab{a}}).
\newblock \bibinfo{title}{Progressive differentiable architecture search:
  Bridging the depth gap between search and evaluation}.
\newblock In {\it \bibinfo{booktitle}{Proceedings of the IEEE International
  Conference on Computer Vision}\/} (pp. \bibinfo{pages}{1294--1303}).
\bibitem[{Chen et~al.(2019{\natexlab{b}})Chen, Xie, Wu \& Tian}]{pdarts}
\bibinfo{author}{Chen, X.}, \bibinfo{author}{Xie, L.}, \bibinfo{author}{Wu,
  J.}, \& \bibinfo{author}{Tian, Q.} (\bibinfo{year}{2019}{\natexlab{b}}).
\newblock \bibinfo{title}{Progressive differentiable architecture search:
  Bridging the depth gap between search and evaluation}.
\newblock {\it \bibinfo{journal}{In Proceedings of the IEEE International
  Conference on Computer Vision}\/},  (pp. \bibinfo{pages}{1294--1303}).
\bibitem[{Davison et~al.(2016)Davison, Lansley, Costen, Tan \&
  Yap}]{davison2016samm}
\bibinfo{author}{Davison, A.~K.}, \bibinfo{author}{Lansley, C.},
  \bibinfo{author}{Costen, N.}, \bibinfo{author}{Tan, K.}, \&
  \bibinfo{author}{Yap, M.~H.} (\bibinfo{year}{2016}).
\newblock \bibinfo{title}{Samm: A spontaneous micro-facial movement dataset}.
\newblock {\it \bibinfo{journal}{IEEE Transactions on Affective Computing}\/},
  {\it \bibinfo{volume}{9}\/}, \bibinfo{pages}{116--129}.
\bibitem[{Ekman(2003)}]{ekman2003darwin}
\bibinfo{author}{Ekman, P.} (\bibinfo{year}{2003}).
\newblock \bibinfo{title}{Darwin, deception, and facial expression}.
\newblock {\it \bibinfo{journal}{Annals of the New York Academy of
  Sciences}\/},  {\it \bibinfo{volume}{1000}\/}, \bibinfo{pages}{205--221}.
\bibitem[{Ekman \& Friesen(1971)}]{ekman1971constants}
\bibinfo{author}{Ekman, P.}, \& \bibinfo{author}{Friesen, W.~V.}
  (\bibinfo{year}{1971}).
\newblock \bibinfo{title}{Constants across cultures in the face and emotion.}
\newblock {\it \bibinfo{journal}{Journal of personality and social
  psychology}\/},  {\it \bibinfo{volume}{17}\/}, \bibinfo{pages}{124}.
\bibitem[{Fan et~al.(2018)Fan, Lam \& Li}]{fan2018multi}
\bibinfo{author}{Fan, Y.}, \bibinfo{author}{Lam, J.~C.}, \&
  \bibinfo{author}{Li, V.~O.} (\bibinfo{year}{2018}).
\newblock \bibinfo{title}{Multi-region ensemble convolutional neural network
  for facial expression recognition}.
\newblock In {\it \bibinfo{booktitle}{International Conference on Artificial
  Neural Networks}\/} (pp. \bibinfo{pages}{84--94}).
\newblock \bibinfo{organization}{Springer}.
\bibitem[{Fan et~al.(2020)Fan, Li \& Lam}]{fan2020facial}
\bibinfo{author}{Fan, Y.}, \bibinfo{author}{Li, V.}, \& \bibinfo{author}{Lam,
  J.~C.} (\bibinfo{year}{2020}).
\newblock \bibinfo{title}{Facial expression recognition with deeply-supervised
  attention network}.
\newblock {\it \bibinfo{journal}{IEEE Transactions on Affective Computing}\/},
  .
\bibitem[{Gan et~al.(2019)Gan, Liong, Yau, Huang \& Tan}]{gan2019off}
\bibinfo{author}{Gan, Y.}, \bibinfo{author}{Liong, S.-T.},
  \bibinfo{author}{Yau, W.-C.}, \bibinfo{author}{Huang, Y.-C.}, \&
  \bibinfo{author}{Tan, L.-K.} (\bibinfo{year}{2019}).
\newblock \bibinfo{title}{Off-apexnet on micro-expression recognition system}.
\newblock {\it \bibinfo{journal}{Signal Processing: Image Communication}\/},
  {\it \bibinfo{volume}{74}\/}, \bibinfo{pages}{129--139}.
\bibitem[{Georgescu et~al.(2018)Georgescu, Ionescu \&
  Popescu}]{Georgescu2018LocalLW}
\bibinfo{author}{Georgescu, M.-I.}, \bibinfo{author}{Ionescu, R.~T.}, \&
  \bibinfo{author}{Popescu, M.} (\bibinfo{year}{2018}).
\newblock \bibinfo{title}{Local learning with deep and handcrafted features for
  facial expression recognition}.
\newblock {\it \bibinfo{journal}{IEEE Access}\/},  {\it \bibinfo{volume}{7}\/},
  \bibinfo{pages}{64827--64836}.
\bibitem[{Giannopoulos et~al.(2018)Giannopoulos, Perikos \&
  Hatzilygeroudis}]{Giannopoulos2018DeepLA}
\bibinfo{author}{Giannopoulos, P.}, \bibinfo{author}{Perikos, I.}, \&
  \bibinfo{author}{Hatzilygeroudis, I.} (\bibinfo{year}{2018}).
\newblock \bibinfo{title}{Deep learning approaches for facial emotion
  recognition: A case study on fer-2013}.
\newblock In {\it \bibinfo{booktitle}{Advances in hybridization of intelligent
  methods}\/} (pp. \bibinfo{pages}{1--16}).
\newblock \bibinfo{publisher}{Springer}.
\bibitem[{Guoying \& Pietikainen(2007)}]{Zhao}
\bibinfo{author}{Guoying, Z.}, \& \bibinfo{author}{Pietikainen, M.}
  (\bibinfo{year}{2007}).
\newblock \bibinfo{title}{Dynamic texture recognition using local binary
  patterns with an application to facial expressions}.
\newblock In {\it \bibinfo{booktitle}{IEEE transactions on pattern analysis and
  machine intelligence}\/} (pp. \bibinfo{pages}{915--928}).
\newblock volume~\bibinfo{volume}{29}.
\bibitem[{Happy \& Routray(2017)}]{HappyFHOFH}
\bibinfo{author}{Happy, S.~L.}, \& \bibinfo{author}{Routray, A.}
  (\bibinfo{year}{2017}).
\newblock \bibinfo{title}{Fuzzy histogram of optical flow orientations for
  micro-expression recognition}.
\newblock In {\it \bibinfo{booktitle}{IEEE Transactions on Affective
  Computing}\/}.
\bibitem[{He et~al.(2015)He, Zhang, Ren \& Sun}]{He2015DeepRL}
\bibinfo{author}{He, K.}, \bibinfo{author}{Zhang, X.}, \bibinfo{author}{Ren,
  S.}, \& \bibinfo{author}{Sun, J.} (\bibinfo{year}{2015}).
\newblock \bibinfo{title}{Deep residual learning for image recognition}.
\newblock {\it \bibinfo{journal}{2016 IEEE Conference on Computer Vision and
  Pattern Recognition (CVPR)}\/},  (pp. \bibinfo{pages}{770--778}).
\bibitem[{Howard et~al.(2017)Howard, Zhu, Chen, Kalenichenko, Wang, Weyand,
  Andreetto \& Adam}]{Howard2017MobileNetsEC}
\bibinfo{author}{Howard, A.~G.}, \bibinfo{author}{Zhu, M.},
  \bibinfo{author}{Chen, B.}, \bibinfo{author}{Kalenichenko, D.},
  \bibinfo{author}{Wang, W.}, \bibinfo{author}{Weyand, T.},
  \bibinfo{author}{Andreetto, M.}, \& \bibinfo{author}{Adam, H.}
  (\bibinfo{year}{2017}).
\newblock \bibinfo{title}{Mobilenets: Efficient convolutional neural networks
  for mobile vision applications}.
\newblock {\it \bibinfo{journal}{ArXiv}\/},  {\it
  \bibinfo{volume}{abs/1704.04861}\/}.
\bibitem[{Ionescu et~al.(2013)Ionescu, Popescu \& Grozea}]{Ionescu2013LocalLT}
\bibinfo{author}{Ionescu, R.~T.}, \bibinfo{author}{Popescu, M.}, \&
  \bibinfo{author}{Grozea, C.} (\bibinfo{year}{2013}).
\newblock \bibinfo{title}{Local learning to improve bag of visual words model
  for facial expression recognition}.
\newblock In {\it \bibinfo{booktitle}{Workshop on challenges in representation
  learning, ICML}\/}.
\bibitem[{{Iqbal} et~al.(2020){Iqbal}, {Ryu}, {Ramirez Rivera},
  {Makhmudkhujaev}, {Chae} \& {Bae}}]{rivera-2020}
\bibinfo{author}{{Iqbal}, M. T.~B.}, \bibinfo{author}{{Ryu}, B.},
  \bibinfo{author}{{Ramirez Rivera}, A.}, \bibinfo{author}{{Makhmudkhujaev},
  F.}, \bibinfo{author}{{Chae}, O.}, \& \bibinfo{author}{{Bae}, S.}
  (\bibinfo{year}{2020}).
\newblock \bibinfo{title}{Facial expression recognition with active local shape
  pattern and learned-size block representations}.
\newblock {\it \bibinfo{journal}{IEEE Transactions on Affective Computing}\/},
  (pp. \bibinfo{pages}{1--1}). \DOIprefix\doi{10.1109/TAFFC.2020.2995432}.
\bibitem[{Kaggle.com(2019 (accessed December 3, 2019))}]{fer2013}
\bibinfo{author}{Kaggle.com} (\bibinfo{year}{2019 (accessed December 3,
  2019)}).
\newblock {\it \bibinfo{title}{Challenges in Representation Learning: Facial
  Expression Recognition Challenge}\/}.
\newblock \URLprefix
  \url{https://www.kaggle.com/c/challenges-in-representation-learning-facial-expression-recognition-challenge/}.
\bibitem[{Khor et~al.(2019)Khor, See, Liong, Phan \& Lin}]{khor2019dual}
\bibinfo{author}{Khor, H.-Q.}, \bibinfo{author}{See, J.},
  \bibinfo{author}{Liong, S.-T.}, \bibinfo{author}{Phan, R.~C.}, \&
  \bibinfo{author}{Lin, W.} (\bibinfo{year}{2019}).
\newblock \bibinfo{title}{Dual-stream shallow networks for facial
  micro-expression recognition}.
\newblock In {\it \bibinfo{booktitle}{2019 IEEE International Conference on
  Image Processing (ICIP)}\/} (pp. \bibinfo{pages}{36--40}).
\newblock \bibinfo{organization}{IEEE}.
\bibitem[{Khorrami et~al.(2015)Khorrami, Paine \& Huang}]{Khorrami2015DoDN}
\bibinfo{author}{Khorrami, P.}, \bibinfo{author}{Paine, T.~L.}, \&
  \bibinfo{author}{Huang, T.~S.} (\bibinfo{year}{2015}).
\newblock \bibinfo{title}{Do deep neural networks learn facial action units
  when doing expression recognition?}
\newblock {\it \bibinfo{journal}{2015 IEEE International Conference on Computer
  Vision Workshop (ICCVW)}\/},  (pp. \bibinfo{pages}{19--27}).
\bibitem[{Kim et~al.(2017)Kim, Baddar, Jang \& Ro}]{Kimspatiotemporal}
\bibinfo{author}{Kim, D.~H.}, \bibinfo{author}{Baddar, W.~J.},
  \bibinfo{author}{Jang, J.}, \& \bibinfo{author}{Ro, Y.~M.}
  (\bibinfo{year}{2017}).
\newblock \bibinfo{title}{Multi-objective based spatio-temporal feature
  representation learning robust to expression intensity variations for facial
  expression recognition}.
\newblock In {\it \bibinfo{booktitle}{IEEE Transactions on Affective
  Computing}\/} (pp. \bibinfo{pages}{223--236}).
\newblock volume~\bibinfo{volume}{10}.
\bibitem[{Kim et~al.(2022)Kim, Nam \& Ko}]{kim2022facial}
\bibinfo{author}{Kim, S.}, \bibinfo{author}{Nam, J.}, \& \bibinfo{author}{Ko,
  B.~C.} (\bibinfo{year}{2022}).
\newblock \bibinfo{title}{Facial expression recognition based on squeeze vision
  transformer}.
\newblock {\it \bibinfo{journal}{Sensors}\/},  {\it \bibinfo{volume}{22}\/},
  \bibinfo{pages}{3729}.
\bibitem[{Kuo et~al.(2018)Kuo, Lai \& Sarkis}]{kuo2018compact}
\bibinfo{author}{Kuo, C.-M.}, \bibinfo{author}{Lai, S.-H.}, \&
  \bibinfo{author}{Sarkis, M.} (\bibinfo{year}{2018}).
\newblock \bibinfo{title}{A compact deep learning model for robust facial
  expression recognition}.
\newblock In {\it \bibinfo{booktitle}{Proceedings of the IEEE conference on
  computer vision and pattern recognition workshops}\/} (pp.
  \bibinfo{pages}{2121--2129}).
\bibitem[{Li et~al.(2019{\natexlab{a}})Li, Wang, See \& Liu}]{Li3D-Flow}
\bibinfo{author}{Li, J.}, \bibinfo{author}{Wang, Y.}, \bibinfo{author}{See,
  J.}, \& \bibinfo{author}{Liu, W.} (\bibinfo{year}{2019}{\natexlab{a}}).
\newblock \bibinfo{title}{Micro-expression recognition based on 3d flow
  convolutional neural network}.
\newblock In {\it \bibinfo{booktitle}{Pattern Analysis and Applications}\/}
  (pp. \bibinfo{pages}{1331--1339}).
\newblock volume~\bibinfo{volume}{22}.
\bibitem[{Li \& Deng(2020)}]{li2020deep}
\bibinfo{author}{Li, S.}, \& \bibinfo{author}{Deng, W.} (\bibinfo{year}{2020}).
\newblock \bibinfo{title}{Deep facial expression recognition: A survey}.
\newblock {\it \bibinfo{journal}{IEEE Transactions on Affective Computing}\/},
  .
\bibitem[{{Li} \& {Deng}(2020)}]{lideng-taf}
\bibinfo{author}{{Li}, S.}, \& \bibinfo{author}{{Deng}, W.}
  (\bibinfo{year}{2020}).
\newblock \bibinfo{title}{A deeper look at facial expression dataset bias}.
\newblock {\it \bibinfo{journal}{IEEE Transactions on Affective Computing}\/},
  (pp. \bibinfo{pages}{1--1}). \DOIprefix\doi{10.1109/TAFFC.2020.2973158}.
\bibitem[{Li et~al.(2017)Li, Deng \& Du}]{li2017reliable}
\bibinfo{author}{Li, S.}, \bibinfo{author}{Deng, W.}, \& \bibinfo{author}{Du,
  J.} (\bibinfo{year}{2017}).
\newblock \bibinfo{title}{Reliable crowdsourcing and deep locality-preserving
  learning for expression recognition in the wild}.
\newblock In {\it \bibinfo{booktitle}{Proceedings of the IEEE conference on
  computer vision and pattern recognition}\/} (pp.
  \bibinfo{pages}{2852--2861}).
\bibitem[{Li et~al.(2021)Li, Li, Wen, Shi, Yang, Zhou \& Huang}]{li2021auto}
\bibinfo{author}{Li, S.}, \bibinfo{author}{Li, W.}, \bibinfo{author}{Wen, S.},
  \bibinfo{author}{Shi, K.}, \bibinfo{author}{Yang, Y.}, \bibinfo{author}{Zhou,
  P.}, \& \bibinfo{author}{Huang, T.} (\bibinfo{year}{2021}).
\newblock \bibinfo{title}{Auto-fernet: A facial expression recognition network
  with architecture search}.
\newblock {\it \bibinfo{journal}{IEEE Transactions on Network Science and
  Engineering}\/},  {\it \bibinfo{volume}{8}\/}, \bibinfo{pages}{2213--2222}.
\bibitem[{Li et~al.(2013)Li, Pfister, Huang, Zhao \&
  Pietik{\"a}inen}]{li2013spontaneous}
\bibinfo{author}{Li, X.}, \bibinfo{author}{Pfister, T.},
  \bibinfo{author}{Huang, X.}, \bibinfo{author}{Zhao, G.}, \&
  \bibinfo{author}{Pietik{\"a}inen, M.} (\bibinfo{year}{2013}).
\newblock \bibinfo{title}{A spontaneous micro-expression database: Inducement,
  collection and baseline}.
\newblock In {\it \bibinfo{booktitle}{2013 10th IEEE International Conference
  and Workshops on Automatic face and gesture recognition (fg)}\/} (pp.
  \bibinfo{pages}{1--6}).
\newblock \bibinfo{organization}{IEEE}.
\bibitem[{Li et~al.(2020{\natexlab{a}})Li, Huang \& Zhao}]{li2020joint}
\bibinfo{author}{Li, Y.}, \bibinfo{author}{Huang, X.}, \&
  \bibinfo{author}{Zhao, G.} (\bibinfo{year}{2020}{\natexlab{a}}).
\newblock \bibinfo{title}{Joint local and global information learning with
  single apex frame detection for micro-expression recognition}.
\newblock {\it \bibinfo{journal}{IEEE Transactions on Image Processing}\/},
  {\it \bibinfo{volume}{30}\/}, \bibinfo{pages}{249--263}.
\bibitem[{Li et~al.(2020{\natexlab{b}})Li, Lu, Li, Zhang \&
  Zhang}]{li2020facial}
\bibinfo{author}{Li, Y.}, \bibinfo{author}{Lu, G.}, \bibinfo{author}{Li, J.},
  \bibinfo{author}{Zhang, Z.}, \& \bibinfo{author}{Zhang, D.}
  (\bibinfo{year}{2020}{\natexlab{b}}).
\newblock \bibinfo{title}{Facial expression recognition in the wild using
  multi-level features and attention mechanisms}.
\newblock {\it \bibinfo{journal}{IEEE Transactions on Affective Computing}\/},
  .
\bibitem[{Li et~al.(2019{\natexlab{b}})Li, Zeng, Shan \&
  Chen}]{li2018occlusion}
\bibinfo{author}{Li, Y.}, \bibinfo{author}{Zeng, J.}, \bibinfo{author}{Shan,
  S.}, \& \bibinfo{author}{Chen, X.} (\bibinfo{year}{2019}{\natexlab{b}}).
\newblock \bibinfo{title}{Occlusion aware facial expression recognition using
  cnn with attention mechanism}.
\newblock {\it \bibinfo{journal}{IEEE Transactions on Image Processing}\/},
  {\it \bibinfo{volume}{28}\/}, \bibinfo{pages}{2439--2450}.
\bibitem[{Liong et~al.(2019{\natexlab{a}})Liong, Gan, See, Khor \&
  Huang}]{liong2019shallow}
\bibinfo{author}{Liong, S.-T.}, \bibinfo{author}{Gan, Y.},
  \bibinfo{author}{See, J.}, \bibinfo{author}{Khor, H.-Q.}, \&
  \bibinfo{author}{Huang, Y.-C.} (\bibinfo{year}{2019}{\natexlab{a}}).
\newblock \bibinfo{title}{Shallow triple stream three-dimensional cnn (ststnet)
  for micro-expression recognition}.
\newblock In {\it \bibinfo{booktitle}{2019 14th IEEE International Conference
  on Automatic Face \& Gesture Recognition (FG 2019)}\/} (pp.
  \bibinfo{pages}{1--5}).
\newblock \bibinfo{organization}{IEEE}.
\bibitem[{Liong et~al.(2019{\natexlab{b}})Liong, Gan, John~See \&
  Huang}]{ststnet}
\bibinfo{author}{Liong, S.-T.}, \bibinfo{author}{Gan, Y.~S.},
  \bibinfo{author}{John~See, H.-Q.~K.}, \& \bibinfo{author}{Huang, Y.-C.}
  (\bibinfo{year}{2019}{\natexlab{b}}).
\newblock \bibinfo{title}{Shallow triple stream three-dimensional cnn (ststnet)
  for micro-expression recognition}.
\newblock In {\it \bibinfo{booktitle}{In 2019 14th IEEE International
  Conference on Automatic Face \& Gesture Recognition}\/} (pp.
  \bibinfo{pages}{1--5}).
\newblock \bibinfo{publisher}{IEEE}.
\bibitem[{Liu et~al.(2019{\natexlab{a}})Liu, Chen, Schroff, Adam, Hua, Yuille
  \& Fei-Fei}]{liu2019auto}
\bibinfo{author}{Liu, C.}, \bibinfo{author}{Chen, L.-C.},
  \bibinfo{author}{Schroff, F.}, \bibinfo{author}{Adam, H.},
  \bibinfo{author}{Hua, W.}, \bibinfo{author}{Yuille, A.~L.}, \&
  \bibinfo{author}{Fei-Fei, L.} (\bibinfo{year}{2019}{\natexlab{a}}).
\newblock \bibinfo{title}{Auto-deeplab: Hierarchical neural architecture search
  for semantic image segmentation}.
\newblock In {\it \bibinfo{booktitle}{Proceedings of the IEEE Conference on
  Computer Vision and Pattern Recognition}\/} (pp. \bibinfo{pages}{82--92}).
\bibitem[{Liu et~al.(2018{\natexlab{a}})Liu, Zoph, Neumann, Shlens, Hua, Li,
  Fei-Fei, Yuille, Huang \& Murphy}]{liu2018progressive}
\bibinfo{author}{Liu, C.}, \bibinfo{author}{Zoph, B.},
  \bibinfo{author}{Neumann, M.}, \bibinfo{author}{Shlens, J.},
  \bibinfo{author}{Hua, W.}, \bibinfo{author}{Li, L.-J.},
  \bibinfo{author}{Fei-Fei, L.}, \bibinfo{author}{Yuille, A.},
  \bibinfo{author}{Huang, J.}, \& \bibinfo{author}{Murphy, K.}
  (\bibinfo{year}{2018}{\natexlab{a}}).
\newblock \bibinfo{title}{Progressive neural architecture search}.
\newblock In {\it \bibinfo{booktitle}{Proceedings of the European Conference on
  Computer Vision (ECCV)}\/} (pp. \bibinfo{pages}{19--34}).
\bibitem[{Liu et~al.(2018{\natexlab{b}})Liu, Simonyan \& Yang}]{liu2018darts}
\bibinfo{author}{Liu, H.}, \bibinfo{author}{Simonyan, K.}, \&
  \bibinfo{author}{Yang, Y.} (\bibinfo{year}{2018}{\natexlab{b}}).
\newblock \bibinfo{title}{Darts: Differentiable architecture search}.
\newblock {\it \bibinfo{journal}{arXiv preprint arXiv:1806.09055}\/}, .
\bibitem[{Liu et~al.(2016)Liu, Zhang \& Pan}]{Liu2016FacialER}
\bibinfo{author}{Liu, K.}, \bibinfo{author}{Zhang, M.}, \&
  \bibinfo{author}{Pan, Z.} (\bibinfo{year}{2016}).
\newblock \bibinfo{title}{Facial expression recognition with cnn ensemble}.
\newblock {\it \bibinfo{journal}{2016 International Conference on Cyberworlds
  (CW)}\/},  (pp. \bibinfo{pages}{163--166}).
\bibitem[{Liu et~al.(2014)Liu, Han, Meng \& Tong}]{Liu2014FacialER}
\bibinfo{author}{Liu, P.}, \bibinfo{author}{Han, S.}, \bibinfo{author}{Meng,
  Z.}, \& \bibinfo{author}{Tong, Y.} (\bibinfo{year}{2014}).
\newblock \bibinfo{title}{Facial expression recognition via a boosted deep
  belief network}.
\newblock {\it \bibinfo{journal}{2014 IEEE Conference on Computer Vision and
  Pattern Recognition}\/},  (pp. \bibinfo{pages}{1805--1812}).
\bibitem[{Liu et~al.(2019{\natexlab{b}})Liu, Du, Zheng \&
  Gedeon}]{liu2019neural}
\bibinfo{author}{Liu, Y.}, \bibinfo{author}{Du, H.}, \bibinfo{author}{Zheng,
  L.}, \& \bibinfo{author}{Gedeon, T.} (\bibinfo{year}{2019}{\natexlab{b}}).
\newblock \bibinfo{title}{A neural micro-expression recognizer}.
\newblock In {\it \bibinfo{booktitle}{2019 14th IEEE international conference
  on automatic face \& gesture recognition (FG 2019)}\/} (pp.
  \bibinfo{pages}{1--4}).
\newblock \bibinfo{organization}{IEEE}.
\bibitem[{Liu et~al.(2018{\natexlab{c}})Liu, Li,  \& Lai}]{sparseMDMO}
\bibinfo{author}{Liu, Y.-J.}, \bibinfo{author}{Li, B.-J.}, , \&
  \bibinfo{author}{Lai, Y.-K.} (\bibinfo{year}{2018}{\natexlab{c}}).
\newblock \bibinfo{title}{Sparse mdmo: Learning a discriminative feature for
  spontaneous micro-expression recognition}.
\newblock In {\it \bibinfo{booktitle}{IEEE Transactions on Affective
  Computing}\/}.
\bibitem[{Lopes et~al.(2017)Lopes, de~Aguiar, de~Souza \&
  Oliveira-Santos}]{Lopes2017FacialER}
\bibinfo{author}{Lopes, A.~T.}, \bibinfo{author}{de~Aguiar, E.},
  \bibinfo{author}{de~Souza, A.~F.}, \& \bibinfo{author}{Oliveira-Santos, T.}
  (\bibinfo{year}{2017}).
\newblock \bibinfo{title}{Facial expression recognition with convolutional
  neural networks: Coping with few data and the training sample order}.
\newblock {\it \bibinfo{journal}{Pattern Recognition}\/},  {\it
  \bibinfo{volume}{61}\/}, \bibinfo{pages}{610--628}.
\bibitem[{Lucey et~al.(2010)Lucey, Cohn, Kanade, Saragih, Ambadar \&
  Matthews}]{lucey2010extended}
\bibinfo{author}{Lucey, P.}, \bibinfo{author}{Cohn, J.~F.},
  \bibinfo{author}{Kanade, T.}, \bibinfo{author}{Saragih, J.},
  \bibinfo{author}{Ambadar, Z.}, \& \bibinfo{author}{Matthews, I.}
  (\bibinfo{year}{2010}).
\newblock \bibinfo{title}{The extended cohn-kanade dataset (ck+): A complete
  dataset for action unit and emotion-specified expression}.
\newblock In {\it \bibinfo{booktitle}{2010 ieee computer society conference on
  computer vision and pattern recognition-workshops}\/} (pp.
  \bibinfo{pages}{94--101}).
\newblock \bibinfo{organization}{IEEE}.
\bibitem[{Mandal et~al.(2019)Mandal, Verma, Mathur, Vipparthi, Murala \&
  Deveerasetty}]{Mandal2019RegionalAA}
\bibinfo{author}{Mandal, M.}, \bibinfo{author}{Verma, M.},
  \bibinfo{author}{Mathur, S.}, \bibinfo{author}{Vipparthi, S.~K.},
  \bibinfo{author}{Murala, S.}, \& \bibinfo{author}{Deveerasetty, K.~K.}
  (\bibinfo{year}{2019}).
\newblock \bibinfo{title}{Regional adaptive affinitive patterns (radap) with
  logical operators for facial expression recognition}.
\newblock {\it \bibinfo{journal}{IET Image Processing}\/},  {\it
  \bibinfo{volume}{13}\/}, \bibinfo{pages}{850--861}.
\bibitem[{Mavadati et~al.(2013)Mavadati, Mahoor, Bartlett, Trinh \&
  Cohn}]{mavadati2013disfa}
\bibinfo{author}{Mavadati, S.~M.}, \bibinfo{author}{Mahoor, M.~H.},
  \bibinfo{author}{Bartlett, K.}, \bibinfo{author}{Trinh, P.}, \&
  \bibinfo{author}{Cohn, J.~F.} (\bibinfo{year}{2013}).
\newblock \bibinfo{title}{Disfa: A spontaneous facial action intensity
  database}.
\newblock {\it \bibinfo{journal}{IEEE Transactions on Affective Computing}\/},
  {\it \bibinfo{volume}{4}\/}, \bibinfo{pages}{151--160}.
\bibitem[{Mollahosseini et~al.(2016)Mollahosseini, Chan \&
  Mahoor}]{Mollahosseini2016GoingDI}
\bibinfo{author}{Mollahosseini, A.}, \bibinfo{author}{Chan, D.}, \&
  \bibinfo{author}{Mahoor, M.~H.} (\bibinfo{year}{2016}).
\newblock \bibinfo{title}{Going deeper in facial expression recognition using
  deep neural networks}.
\newblock {\it \bibinfo{journal}{2016 IEEE Winter Conference on Applications of
  Computer Vision (WACV)}\/},  (pp. \bibinfo{pages}{1--10}).
\bibitem[{Negrinho \& Gordon(2017)}]{negrinho2017deeparchitect}
\bibinfo{author}{Negrinho, R.}, \& \bibinfo{author}{Gordon, G.}
  (\bibinfo{year}{2017}).
\newblock \bibinfo{title}{Deeparchitect: Automatically designing and training
  deep architectures}.
\newblock {\it \bibinfo{journal}{arXiv preprint arXiv:1704.08792}\/}, .
\bibitem[{Pasupa \& Sunhem(2016)}]{pasupa2016comparison}
\bibinfo{author}{Pasupa, K.}, \& \bibinfo{author}{Sunhem, W.}
  (\bibinfo{year}{2016}).
\newblock \bibinfo{title}{A comparison between shallow and deep architecture
  classifiers on small dataset}.
\newblock In {\it \bibinfo{booktitle}{2016 8th International Conference on
  Information Technology and Electrical Engineering (ICITEE)}\/} (pp.
  \bibinfo{pages}{1--6}).
\newblock \bibinfo{organization}{IEEE}.
\bibitem[{Pons \& Masip(2017)}]{pons2017supervised}
\bibinfo{author}{Pons, G.}, \& \bibinfo{author}{Masip, D.}
  (\bibinfo{year}{2017}).
\newblock \bibinfo{title}{Supervised committee of convolutional neural networks
  in automated facial expression analysis}.
\newblock {\it \bibinfo{journal}{IEEE Transactions on Affective Computing}\/},
  {\it \bibinfo{volume}{9}\/}, \bibinfo{pages}{343--350}.
\bibitem[{Qu et~al.(2017)Qu, Wang, Yan, Li, Wu \& Fu}]{qu2017cas}
\bibinfo{author}{Qu, F.}, \bibinfo{author}{Wang, S.-J.}, \bibinfo{author}{Yan,
  W.-J.}, \bibinfo{author}{Li, H.}, \bibinfo{author}{Wu, S.}, \&
  \bibinfo{author}{Fu, X.} (\bibinfo{year}{2017}).
\newblock \bibinfo{title}{Cas(me)\^2: A database for spontaneous
  macro-expression and micro-expression spotting and recognition}.
\newblock {\it \bibinfo{journal}{IEEE Transactions on Affective Computing}\/},
  {\it \bibinfo{volume}{9}\/}, \bibinfo{pages}{424--436}.
\bibitem[{Real et~al.(2019)Real, Aggarwal, Huang \& Le}]{real2019regularized}
\bibinfo{author}{Real, E.}, \bibinfo{author}{Aggarwal, A.},
  \bibinfo{author}{Huang, Y.}, \& \bibinfo{author}{Le, Q.~V.}
  (\bibinfo{year}{2019}).
\newblock \bibinfo{title}{Regularized evolution for image classifier
  architecture search}.
\newblock In {\it \bibinfo{booktitle}{Proceedings of the aaai conference on
  artificial intelligence}\/} (pp. \bibinfo{pages}{4780--4789}).
\newblock volume~\bibinfo{volume}{33}.
\bibitem[{{Reddy Sai Prasanna Teja} et~al.(2019){Reddy Sai Prasanna Teja},
  {Surya Teja Karri}, {Shiv Ram Dubey} \& {Snehasis Mukherjee}}]{ReddyFuse}
\bibinfo{author}{{Reddy Sai Prasanna Teja}}, \bibinfo{author}{{Surya Teja
  Karri}}, \bibinfo{author}{{Shiv Ram Dubey}}, \& \bibinfo{author}{{Snehasis
  Mukherjee}} (\bibinfo{year}{2019}).
\newblock \bibinfo{title}{Spontaneous facial micro-expression recognition using
  3d spatiotemporal convolutional neural networks}.
\newblock In {\it \bibinfo{booktitle}{In 2019 International Joint Conference on
  Neural Networks (IJCNN)}\/} (pp. \bibinfo{pages}{1--8}).
\newblock \bibinfo{publisher}{IEEE}.
\bibitem[{{S. L. Happy} et~al.(2017){S. L. Happy}, {Priyadarshi Patnaik},
  {Aurobinda Routray} \& {Rajlakshmi Guha}}]{Happy2017TheIS}
\bibinfo{author}{{S. L. Happy}}, \bibinfo{author}{{Priyadarshi Patnaik}},
  \bibinfo{author}{{Aurobinda Routray}}, \& \bibinfo{author}{{Rajlakshmi Guha}}
  (\bibinfo{year}{2017}).
\newblock \bibinfo{title}{The indian spontaneous expression database for
  emotion recognition}.
\newblock {\it \bibinfo{journal}{IEEE Transactions on Affective Computing}\/},
  {\it \bibinfo{volume}{8}\/}, \bibinfo{pages}{131--142}.
\bibitem[{See et~al.(2019)See, Yap, Li, Hong \& Wang}]{see2019megc}
\bibinfo{author}{See, J.}, \bibinfo{author}{Yap, M.~H.}, \bibinfo{author}{Li,
  J.}, \bibinfo{author}{Hong, X.}, \& \bibinfo{author}{Wang, S.-J.}
  (\bibinfo{year}{2019}).
\newblock \bibinfo{title}{Megc 2019--the second facial micro-expressions grand
  challenge}.
\newblock In {\it \bibinfo{booktitle}{2019 14th IEEE International Conference
  on Automatic Face \& Gesture Recognition (FG 2019)}\/} (pp.
  \bibinfo{pages}{1--5}).
\newblock \bibinfo{organization}{IEEE}.
\bibitem[{Simonyan \& Zisserman(2014)}]{Simonyan2014VeryDC}
\bibinfo{author}{Simonyan, K.}, \& \bibinfo{author}{Zisserman, A.}
  (\bibinfo{year}{2014}).
\newblock \bibinfo{title}{Very deep convolutional networks for large-scale
  image recognition}.
\newblock {\it \bibinfo{journal}{CoRR}\/},  {\it
  \bibinfo{volume}{abs/1409.1556}\/}.
\bibitem[{Song et~al.(2019)Song, Li, Zong, Zhu, Zheng, Shi \&
  Zhao}]{song2019recognizing}
\bibinfo{author}{Song, B.}, \bibinfo{author}{Li, K.}, \bibinfo{author}{Zong,
  Y.}, \bibinfo{author}{Zhu, J.}, \bibinfo{author}{Zheng, W.},
  \bibinfo{author}{Shi, J.}, \& \bibinfo{author}{Zhao, L.}
  (\bibinfo{year}{2019}).
\newblock \bibinfo{title}{Recognizing spontaneous micro-expression using a
  three-stream convolutional neural network}.
\newblock {\it \bibinfo{journal}{IEEE Access}\/},  {\it \bibinfo{volume}{7}\/},
  \bibinfo{pages}{184537--184551}.
\bibitem[{Srivastava et~al.(2015)Srivastava, Greff \&
  Schmidhuber}]{srivastava2015training}
\bibinfo{author}{Srivastava, R.~K.}, \bibinfo{author}{Greff, K.}, \&
  \bibinfo{author}{Schmidhuber, J.} (\bibinfo{year}{2015}).
\newblock \bibinfo{title}{Training very deep networks}.
\newblock In {\it \bibinfo{booktitle}{Advances in neural information processing
  systems}\/} (pp. \bibinfo{pages}{2377--2385}).
\bibitem[{{Sun} et~al.(2020){Sun}, {Cao}, {Li}, {He} \& {Yu}}]{suncao-taf}
\bibinfo{author}{{Sun}, B.}, \bibinfo{author}{{Cao}, S.},
  \bibinfo{author}{{Li}, D.}, \bibinfo{author}{{He}, J.}, \&
  \bibinfo{author}{{Yu}, L.} (\bibinfo{year}{2020}).
\newblock \bibinfo{title}{Dynamic micro-expression recognition using knowledge
  distillation}.
\newblock {\it \bibinfo{journal}{IEEE Transactions on Affective Computing}\/},
  (pp. \bibinfo{pages}{1--1}). \DOIprefix\doi{10.1109/TAFFC.2020.2986962}.
\bibitem[{Van~Quang et~al.(2019)Van~Quang, Chun \&
  Tokuyama}]{van2019capsulenet}
\bibinfo{author}{Van~Quang, N.}, \bibinfo{author}{Chun, J.}, \&
  \bibinfo{author}{Tokuyama, T.} (\bibinfo{year}{2019}).
\newblock \bibinfo{title}{Capsulenet for micro-expression recognition}.
\newblock In {\it \bibinfo{booktitle}{2019 14th IEEE International Conference
  on Automatic Face \& Gesture Recognition (FG 2019)}\/} (pp.
  \bibinfo{pages}{1--7}).
\newblock \bibinfo{organization}{IEEE}.
\bibitem[{Verma et~al.(2021)Verma, Reddy, Meedimale, Mandal \&
  Vipparthi}]{verma2021automer}
\bibinfo{author}{Verma, M.}, \bibinfo{author}{Reddy, M. S.~K.},
  \bibinfo{author}{Meedimale, Y.~R.}, \bibinfo{author}{Mandal, M.}, \&
  \bibinfo{author}{Vipparthi, S.~K.} (\bibinfo{year}{2021}).
\newblock \bibinfo{title}{Automer: Spatiotemporal neural architecture search
  for microexpression recognition}.
\newblock {\it \bibinfo{journal}{IEEE Transactions on Neural Networks and
  Learning Systems}\/}, .
\bibitem[{Verma et~al.(2019{\natexlab{a}})Verma, Vipparthi \&
  Singh}]{verma2019hinet}
\bibinfo{author}{Verma, M.}, \bibinfo{author}{Vipparthi, S.~K.}, \&
  \bibinfo{author}{Singh, G.} (\bibinfo{year}{2019}{\natexlab{a}}).
\newblock \bibinfo{title}{Hinet: Hybrid inherited feature learning network for
  facial expression recognition}.
\newblock {\it \bibinfo{journal}{IEEE Letters of the Computer Society}\/},
  {\it \bibinfo{volume}{2}\/}, \bibinfo{pages}{36--39}.
\bibitem[{{Verma} et~al.(2020{\natexlab{a}}){Verma}, {Vipparthi} \&
  {Singh}}]{monu-multimedia}
\bibinfo{author}{{Verma}, M.}, \bibinfo{author}{{Vipparthi}, S.~K.}, \&
  \bibinfo{author}{{Singh}, G.} (\bibinfo{year}{2020}{\natexlab{a}}).
\newblock \bibinfo{title}{Affectivenet: Affective-motion feature learning for
  micro expression recognition}.
\newblock {\it \bibinfo{journal}{IEEE MultiMedia}\/},  (pp.
  \bibinfo{pages}{1--1}). \DOIprefix\doi{10.1109/MMUL.2020.3021659}.
\bibitem[{{Verma} et~al.(2020{\natexlab{b}}){Verma}, {Vipparthi} \&
  {Singh}}]{monu-ijcnn}
\bibinfo{author}{{Verma}, M.}, \bibinfo{author}{{Vipparthi}, S.~K.}, \&
  \bibinfo{author}{{Singh}, G.} (\bibinfo{year}{2020}{\natexlab{b}}).
\newblock \bibinfo{title}{Non-linearities improve originet based on active
  imaging for micro expression recognition}.
\newblock In {\it \bibinfo{booktitle}{2020 International Joint Conference on
  Neural Networks (IJCNN)}\/} (pp. \bibinfo{pages}{1--8}).
\newblock \DOIprefix\doi{10.1109/IJCNN48605.2020.9207718}.
\bibitem[{Verma et~al.(2019{\natexlab{b}})Verma, Vipparthi, Singh \&
  Murala}]{verma2019learnet}
\bibinfo{author}{Verma, M.}, \bibinfo{author}{Vipparthi, S.~K.},
  \bibinfo{author}{Singh, G.}, \& \bibinfo{author}{Murala, S.}
  (\bibinfo{year}{2019}{\natexlab{b}}).
\newblock \bibinfo{title}{Learnet: Dynamic imaging network for micro expression
  recognition}.
\newblock {\it \bibinfo{journal}{IEEE Transactions on Image Processing}\/},
  {\it \bibinfo{volume}{29}\/}, \bibinfo{pages}{1618--1627}.
\bibitem[{Wang et~al.(2018{\natexlab{a}})Wang, Peng, Bi \&
  Chen}]{wang2018micro}
\bibinfo{author}{Wang, C.}, \bibinfo{author}{Peng, M.}, \bibinfo{author}{Bi,
  T.}, \& \bibinfo{author}{Chen, T.} (\bibinfo{year}{2018}{\natexlab{a}}).
\newblock \bibinfo{title}{Micro-attention for micro-expression recognition}.
\newblock {\it \bibinfo{journal}{arXiv preprint arXiv:1811.02360}\/}, .
\bibitem[{Wang et~al.(2020{\natexlab{a}})Wang, Peng, Bi \&
  Chen}]{wang2020micro}
\bibinfo{author}{Wang, C.}, \bibinfo{author}{Peng, M.}, \bibinfo{author}{Bi,
  T.}, \& \bibinfo{author}{Chen, T.} (\bibinfo{year}{2020}{\natexlab{a}}).
\newblock \bibinfo{title}{Micro-attention for micro-expression recognition}.
\newblock {\it \bibinfo{journal}{Neurocomputing}\/},  {\it
  \bibinfo{volume}{410}\/}, \bibinfo{pages}{354--362}.
\bibitem[{Wang et~al.(2020{\natexlab{b}})Wang, Peng, Yang, Lu \&
  Qiao}]{Wang_2020_CVPR}
\bibinfo{author}{Wang, K.}, \bibinfo{author}{Peng, X.}, \bibinfo{author}{Yang,
  J.}, \bibinfo{author}{Lu, S.}, \& \bibinfo{author}{Qiao, Y.}
  (\bibinfo{year}{2020}{\natexlab{b}}).
\newblock \bibinfo{title}{Suppressing uncertainties for large-scale facial
  expression recognition}.
\newblock In {\it \bibinfo{booktitle}{Proceedings of the IEEE/CVF Conference on
  Computer Vision and Pattern Recognition (CVPR)}\/}.
\bibitem[{Wang et~al.(2018{\natexlab{b}})Wang, Li, Liu, Yan, Ou, Huang, Xu \&
  Fu}]{CNN-LSTM}
\bibinfo{author}{Wang, S.-J.}, \bibinfo{author}{Li, B.-J.},
  \bibinfo{author}{Liu, Y.-J.}, \bibinfo{author}{Yan, W.-J.},
  \bibinfo{author}{Ou, X.}, \bibinfo{author}{Huang, X.}, \bibinfo{author}{Xu,
  F.}, \& \bibinfo{author}{Fu, X.} (\bibinfo{year}{2018}{\natexlab{b}}).
\newblock \bibinfo{title}{Micro-expression recognition with small sample size
  by transferring long-term convolutional neural network}.
\newblock In {\it \bibinfo{booktitle}{Neurocomputing 312}\/} (pp.
  \bibinfo{pages}{251--262}).
\bibitem[{Wang et~al.(2019)Wang, Sun, Chen, Cao, Zheng, Xu, Qiu \&
  Fu}]{Wang2019AFF}
\bibinfo{author}{Wang, W.}, \bibinfo{author}{Sun, Q.}, \bibinfo{author}{Chen,
  T.}, \bibinfo{author}{Cao, C.}, \bibinfo{author}{Zheng, Z.},
  \bibinfo{author}{Xu, G.}, \bibinfo{author}{Qiu, H.}, \& \bibinfo{author}{Fu,
  Y.} (\bibinfo{year}{2019}).
\newblock \bibinfo{title}{A fine-grained facial expression database for
  end-to-end multi-pose facial expression recognition}.
\newblock {\it \bibinfo{journal}{ArXiv}\/},  {\it
  \bibinfo{volume}{abs/1907.10838}\/}.
\bibitem[{Wen et~al.(2021)Wen, Lin, Wang \& Xu}]{wen2021distract}
\bibinfo{author}{Wen, Z.}, \bibinfo{author}{Lin, W.}, \bibinfo{author}{Wang,
  T.}, \& \bibinfo{author}{Xu, G.} (\bibinfo{year}{2021}).
\newblock \bibinfo{title}{Distract your attention: multi-head cross attention
  network for facial expression recognition}.
\newblock {\it \bibinfo{journal}{arXiv preprint arXiv:2109.07270}\/}, .
\bibitem[{Xia et~al.(2019)Xia, Liang, Hong \& Feng}]{xia2019cross}
\bibinfo{author}{Xia, Z.}, \bibinfo{author}{Liang, H.}, \bibinfo{author}{Hong,
  X.}, \& \bibinfo{author}{Feng, X.} (\bibinfo{year}{2019}).
\newblock \bibinfo{title}{Cross-database micro-expression recognition with deep
  convolutional networks}.
\newblock In {\it \bibinfo{booktitle}{Proceedings of the 2019 3rd International
  Conference on Biometric Engineering and Applications}\/} (pp.
  \bibinfo{pages}{56--60}).
\bibitem[{Xia et~al.(2020)Xia, Peng, Khor, Feng \& Zhao}]{xia2020revealing}
\bibinfo{author}{Xia, Z.}, \bibinfo{author}{Peng, W.}, \bibinfo{author}{Khor,
  H.-Q.}, \bibinfo{author}{Feng, X.}, \& \bibinfo{author}{Zhao, G.}
  (\bibinfo{year}{2020}).
\newblock \bibinfo{title}{Revealing the invisible with model and data shrinking
  for composite-database micro-expression recognition}.
\newblock {\it \bibinfo{journal}{IEEE Transactions on Image Processing}\/},
  {\it \bibinfo{volume}{29}\/}, \bibinfo{pages}{8590--8605}.
\bibitem[{Xie \& Hu(2019)}]{Xie2019FacialER}
\bibinfo{author}{Xie, S.}, \& \bibinfo{author}{Hu, H.} (\bibinfo{year}{2019}).
\newblock \bibinfo{title}{Facial expression recognition using hierarchical
  features with deep comprehensive multipatches aggregation convolutional
  neural networks}.
\newblock {\it \bibinfo{journal}{IEEE Transactions on Multimedia}\/},  {\it
  \bibinfo{volume}{21}\/}, \bibinfo{pages}{211--220}.
\bibitem[{Xie et~al.(2020)Xie, Chen, Pu, Wu \& Lin}]{xie2020adversarial}
\bibinfo{author}{Xie, Y.}, \bibinfo{author}{Chen, T.}, \bibinfo{author}{Pu,
  T.}, \bibinfo{author}{Wu, H.}, \& \bibinfo{author}{Lin, L.}
  (\bibinfo{year}{2020}).
\newblock \bibinfo{title}{Adversarial graph representation adaptation for
  cross-domain facial expression recognition}.
\newblock In {\it \bibinfo{booktitle}{Proceedings of the 28th ACM international
  conference on Multimedia}\/} (pp. \bibinfo{pages}{1255--1264}).
\bibitem[{Xu et~al.(2017)Xu, Zhang,  \& Wang}]{XuDynamicMap}
\bibinfo{author}{Xu, F.}, \bibinfo{author}{Zhang, J.}, , \&
  \bibinfo{author}{Wang, J.~Z.} (\bibinfo{year}{2017}).
\newblock \bibinfo{title}{Microexpression identification and categorization
  using a facial dynamics map}.
\newblock In {\it \bibinfo{booktitle}{IEEE Transactions on Affective
  Computing}\/} (pp. \bibinfo{pages}{254--267}).
\newblock volume~\bibinfo{volume}{8}.
\bibitem[{Yan et~al.(2014)Yan, Li, Wang, Zhao, Liu, Chen \& Fu}]{yan2014casme}
\bibinfo{author}{Yan, W.-J.}, \bibinfo{author}{Li, X.}, \bibinfo{author}{Wang,
  S.-J.}, \bibinfo{author}{Zhao, G.}, \bibinfo{author}{Liu, Y.-J.},
  \bibinfo{author}{Chen, Y.-H.}, \& \bibinfo{author}{Fu, X.}
  (\bibinfo{year}{2014}).
\newblock \bibinfo{title}{Casme ii: An improved spontaneous micro-expression
  database and the baseline evaluation}.
\newblock {\it \bibinfo{journal}{PloS one}\/},  {\it \bibinfo{volume}{9}\/},
  \bibinfo{pages}{e86041}.
\bibitem[{Yan et~al.(2013)Yan, Wu, Liu, Wang \& Fu}]{yan2013casme}
\bibinfo{author}{Yan, W.-J.}, \bibinfo{author}{Wu, Q.}, \bibinfo{author}{Liu,
  Y.-J.}, \bibinfo{author}{Wang, S.-J.}, \& \bibinfo{author}{Fu, X.}
  (\bibinfo{year}{2013}).
\newblock \bibinfo{title}{Casme database: a dataset of spontaneous
  micro-expressions collected from neutralized faces}.
\newblock In {\it \bibinfo{booktitle}{2013 10th IEEE international conference
  and workshops on automatic face and gesture recognition (FG)}\/} (pp.
  \bibinfo{pages}{1--7}).
\newblock \bibinfo{organization}{IEEE}.
\bibitem[{Yang et~al.(2021)Yang, Cheng, Yang, Zhang \& Li}]{yang2021merta}
\bibinfo{author}{Yang, B.}, \bibinfo{author}{Cheng, J.}, \bibinfo{author}{Yang,
  Y.}, \bibinfo{author}{Zhang, B.}, \& \bibinfo{author}{Li, J.}
  (\bibinfo{year}{2021}).
\newblock \bibinfo{title}{Merta: micro-expression recognition with ternary
  attentions}.
\newblock {\it \bibinfo{journal}{Multimedia Tools and Applications}\/},  {\it
  \bibinfo{volume}{80}\/}, \bibinfo{pages}{1--16}.
\bibitem[{Yang et~al.(2018)Yang, Zhang \& Yin}]{Yang2018IdentityAdaptiveFE}
\bibinfo{author}{Yang, H.}, \bibinfo{author}{Zhang, Z.}, \&
  \bibinfo{author}{Yin, L.} (\bibinfo{year}{2018}).
\newblock \bibinfo{title}{Identity-adaptive facial expression recognition
  through expression regeneration using conditional generative adversarial
  networks}.
\newblock {\it \bibinfo{journal}{2018 13th IEEE International Conference on
  Automatic Face \& Gesture Recognition (FG 2018)}\/},  (pp.
  \bibinfo{pages}{294--301}).
\bibitem[{Yong-Jin et~al.(2015)Yong-Jin, Jin-Kai, Wen-Jing, Su-Jing, Guoying \&
  Xiaolan}]{liu2015main}
\bibinfo{author}{Yong-Jin, L.}, \bibinfo{author}{Jin-Kai, Z.},
  \bibinfo{author}{Wen-Jing, Y.}, \bibinfo{author}{Su-Jing, W.},
  \bibinfo{author}{Guoying, Z.}, \& \bibinfo{author}{Xiaolan, F.}
  (\bibinfo{year}{2015}).
\newblock \bibinfo{title}{A main directional mean optical flow feature for
  spontaneous micro-expression recognition}.
\newblock {\it \bibinfo{journal}{IEEE Transactions on Affective Computing}\/},
  {\it \bibinfo{volume}{7}\/}, \bibinfo{pages}{299--310}.
\bibitem[{Yu et~al.(2020{\natexlab{a}})Yu, Qin, Xu, Zhao, Wang, Lei \&
  Zhao}]{yu2020auto}
\bibinfo{author}{Yu, Z.}, \bibinfo{author}{Qin, Y.}, \bibinfo{author}{Xu, X.},
  \bibinfo{author}{Zhao, C.}, \bibinfo{author}{Wang, Z.}, \bibinfo{author}{Lei,
  Z.}, \& \bibinfo{author}{Zhao, G.} (\bibinfo{year}{2020}{\natexlab{a}}).
\newblock \bibinfo{title}{Auto-fas: Searching lightweight networks for face
  anti-spoofing}.
\newblock In {\it \bibinfo{booktitle}{ICASSP 2020-2020 IEEE International
  Conference on Acoustics, Speech and Signal Processing (ICASSP)}\/} (pp.
  \bibinfo{pages}{996--1000}).
\newblock \bibinfo{organization}{IEEE}.
\bibitem[{Yu et~al.(2020{\natexlab{b}})Yu, Wan, Qin, Li, Li \&
  Zhao}]{yu2020fas}
\bibinfo{author}{Yu, Z.}, \bibinfo{author}{Wan, J.}, \bibinfo{author}{Qin, Y.},
  \bibinfo{author}{Li, X.}, \bibinfo{author}{Li, S.~Z.}, \&
  \bibinfo{author}{Zhao, G.} (\bibinfo{year}{2020}{\natexlab{b}}).
\newblock \bibinfo{title}{Nas-fas: Static-dynamic central difference network
  search for face anti-spoofing}.
\newblock {\it \bibinfo{journal}{arXiv preprint arXiv:2011.02062}\/}, .
\bibitem[{Yu et~al.(2020{\natexlab{c}})Yu, Zhao, Wang, Qin, Su, Li, Zhou \&
  Zhao}]{yu2020searching}
\bibinfo{author}{Yu, Z.}, \bibinfo{author}{Zhao, C.}, \bibinfo{author}{Wang,
  Z.}, \bibinfo{author}{Qin, Y.}, \bibinfo{author}{Su, Z.},
  \bibinfo{author}{Li, X.}, \bibinfo{author}{Zhou, F.}, \&
  \bibinfo{author}{Zhao, G.} (\bibinfo{year}{2020}{\natexlab{c}}).
\newblock \bibinfo{title}{Searching central difference convolutional networks
  for face anti-spoofing}.
\newblock In {\it \bibinfo{booktitle}{Proceedings of the IEEE/CVF Conference on
  Computer Vision and Pattern Recognition}\/} (pp.
  \bibinfo{pages}{5295--5305}).
\bibitem[{Zhang et~al.(2016)Zhang, Zheng, Cui, Zong, Yan \&
  Yan}]{zhang2016deep}
\bibinfo{author}{Zhang, T.}, \bibinfo{author}{Zheng, W.}, \bibinfo{author}{Cui,
  Z.}, \bibinfo{author}{Zong, Y.}, \bibinfo{author}{Yan, J.}, \&
  \bibinfo{author}{Yan, K.} (\bibinfo{year}{2016}).
\newblock \bibinfo{title}{A deep neural network-driven feature learning method
  for multi-view facial expression recognition}.
\newblock {\it \bibinfo{journal}{IEEE Transactions on Multimedia}\/},  {\it
  \bibinfo{volume}{18}\/}, \bibinfo{pages}{2528--2536}.
\bibitem[{{Zhang} et~al.(2020){Zhang}, {Zong}, {Zheng}, {Chen}, {Hong}, {Tang},
  {Cui} \& {Zhao}}]{zhang-tkde}
\bibinfo{author}{{Zhang}, T.}, \bibinfo{author}{{Zong}, Y.},
  \bibinfo{author}{{Zheng}, W.}, \bibinfo{author}{{Chen}, C. L.~P.},
  \bibinfo{author}{{Hong}, X.}, \bibinfo{author}{{Tang}, C.},
  \bibinfo{author}{{Cui}, Z.}, \& \bibinfo{author}{{Zhao}, G.}
  (\bibinfo{year}{2020}).
\newblock \bibinfo{title}{Cross-database micro-expression recognition: A
  benchmark}.
\newblock {\it \bibinfo{journal}{IEEE Transactions on Knowledge and Data
  Engineering}\/},  (pp. \bibinfo{pages}{1--1}).
  \DOIprefix\doi{10.1109/TKDE.2020.2985365}.
\bibitem[{Zhao et~al.(2011)Zhao, Huang, Taini, Li \&
  Pietik{\"a}Inen}]{zhao2011facial}
\bibinfo{author}{Zhao, G.}, \bibinfo{author}{Huang, X.},
  \bibinfo{author}{Taini, M.}, \bibinfo{author}{Li, S.~Z.}, \&
  \bibinfo{author}{Pietik{\"a}Inen, M.} (\bibinfo{year}{2011}).
\newblock \bibinfo{title}{Facial expression recognition from near-infrared
  videos}.
\newblock {\it \bibinfo{journal}{Image and Vision Computing}\/},  {\it
  \bibinfo{volume}{29}\/}, \bibinfo{pages}{607--619}.
\bibitem[{Zhao et~al.(2018)Zhao, Cai, Liu, Zhang \& Chen}]{zhao2018feature}
\bibinfo{author}{Zhao, S.}, \bibinfo{author}{Cai, H.}, \bibinfo{author}{Liu,
  H.}, \bibinfo{author}{Zhang, J.}, \& \bibinfo{author}{Chen, S.}
  (\bibinfo{year}{2018}).
\newblock \bibinfo{title}{Feature selection mechanism in cnns for facial
  expression recognition.}
\newblock In {\it \bibinfo{booktitle}{BMVC}\/} (p. \bibinfo{pages}{317}).
\bibitem[{Zheng et~al.(2022)Zheng, Mendieta \& Chen}]{zheng2022poster}
\bibinfo{author}{Zheng, C.}, \bibinfo{author}{Mendieta, M.}, \&
  \bibinfo{author}{Chen, C.} (\bibinfo{year}{2022}).
\newblock \bibinfo{title}{Poster: A pyramid cross-fusion transformer network
  for facial expression recognition}.
\newblock {\it \bibinfo{journal}{arXiv preprint arXiv:2204.04083}\/}, .
\bibitem[{Zhou et~al.(2019)Zhou, Mao \& Xue}]{zhou2019dual}
\bibinfo{author}{Zhou, L.}, \bibinfo{author}{Mao, Q.}, \& \bibinfo{author}{Xue,
  L.} (\bibinfo{year}{2019}).
\newblock \bibinfo{title}{Dual-inception network for cross-database
  micro-expression recognition}.
\newblock In {\it \bibinfo{booktitle}{2019 14th IEEE International Conference
  on Automatic Face \& Gesture Recognition (FG 2019)}\/} (pp.
  \bibinfo{pages}{1--5}).
\newblock \bibinfo{organization}{IEEE}.
\bibitem[{Zhou et~al.(2020)Zhou, Shao \& Mao}]{zhou2020survey}
\bibinfo{author}{Zhou, L.}, \bibinfo{author}{Shao, X.}, \&
  \bibinfo{author}{Mao, Q.} (\bibinfo{year}{2020}).
\newblock \bibinfo{title}{A survey of micro-expression recognition}.
\newblock {\it \bibinfo{journal}{Image and Vision Computing}\/},  (p.
  \bibinfo{pages}{104043}).
\bibitem[{Zoph et~al.(2018)Zoph, Vasudevan, Shlens \& Le}]{zoph2018learning}
\bibinfo{author}{Zoph, B.}, \bibinfo{author}{Vasudevan, V.},
  \bibinfo{author}{Shlens, J.}, \& \bibinfo{author}{Le, Q.~V.}
  (\bibinfo{year}{2018}).
\newblock \bibinfo{title}{Learning transferable architectures for scalable
  image recognition}.
\newblock In {\it \bibinfo{booktitle}{Proceedings of the IEEE conference on
  computer vision and pattern recognition}\/} (pp.
  \bibinfo{pages}{8697--8710}).

\end{thebibliography}

\end{document}